\documentclass[10pt,twocolumn,letterpaper]{article}

\usepackage[pagenumbers]{cvpr}      
\usepackage{multirow}

\definecolor{cvprblue}{rgb}{0.21,0.49,0.74}
\usepackage[pagebackref,breaklinks,colorlinks,allcolors=cvprblue]{hyperref}

\def\paperID{5444} 
\def\confName{CVPR}
\def\confYear{2026}

\title{Clair Obscur: an Illumination-Aware Method for Real-World Image Vectorization}

\author{
    Xingyue Lin, Shuai Peng, Xiangyu Xie, Jianhua Zhu, Yuxuan Zhou, 
    Liangcai Gao\thanks{*Corresponding author} 
    \\
    Wangxuan Institute of Computer Technology, Peking University, Beijing, China \\
    {\tt\small linxingyue23@stu.pku.edu.cn, \{pengshuaipku, xxy, zhujianhuapku, sherco, gaoliangcai\}@pku.edu.cn}
}

\begin{document}
\maketitle
\begin{abstract}
Image vectorization aims to convert raster images into editable, scalable vector representations while preserving visual fidelity. 
Existing vectorization methods struggle to represent complex real-world images, often producing fragmented shapes at the cost of semantic conciseness. In this paper, we propose COVec, an illumination-aware vectorization framework inspired by the Clair-Obscur principle of light–shade contrast. COVec is the first to introduce intrinsic image decomposition in the vector domain, separating an image into albedo, shade, and light layers in a unified vector representation. A semantic-guided initialization and two-stage optimization refine these layers with differentiable rendering. Experiments on various datasets demonstrate that COVec achieves higher visual fidelity and significantly improved editability compared to existing methods.
The code will be released at \url{https://github.com/decade-de/COVec}.
\end{abstract}

\section{Introduction}
\label{sec:intro}

Scalable Vector Graphics (SVG) represent images through geometric primitives, providing a compact and structured description of visual content~\cite{ferraiolo2000scalable}. 
Transform raster images into vector graphics, namely image vectorization, enables scalable, resolution-independent representations that support downstream tasks such as editing, animation, compression, and high-quality rendering across diverse display settings.

While existing vectorization methods perform well on simple images such as icons and emojis, extending them to real-world photographs remains challenging. To match complex tonal variations, current methods~\cite{ma2022towards, zhao2025less, hirschorn2024optimize, wang2025layered, zhou2024segmentation} often produce numerous fragmented shapes.
This fragmentation undermines semantic conciseness\textemdash the ability to represent an image using a small number of coherent, semantically meaningful vector shapes, which in turn limits the editability of the resulting SVG.

\begin{figure}[htbp]
  \centering
   \includegraphics[width=0.85\linewidth]{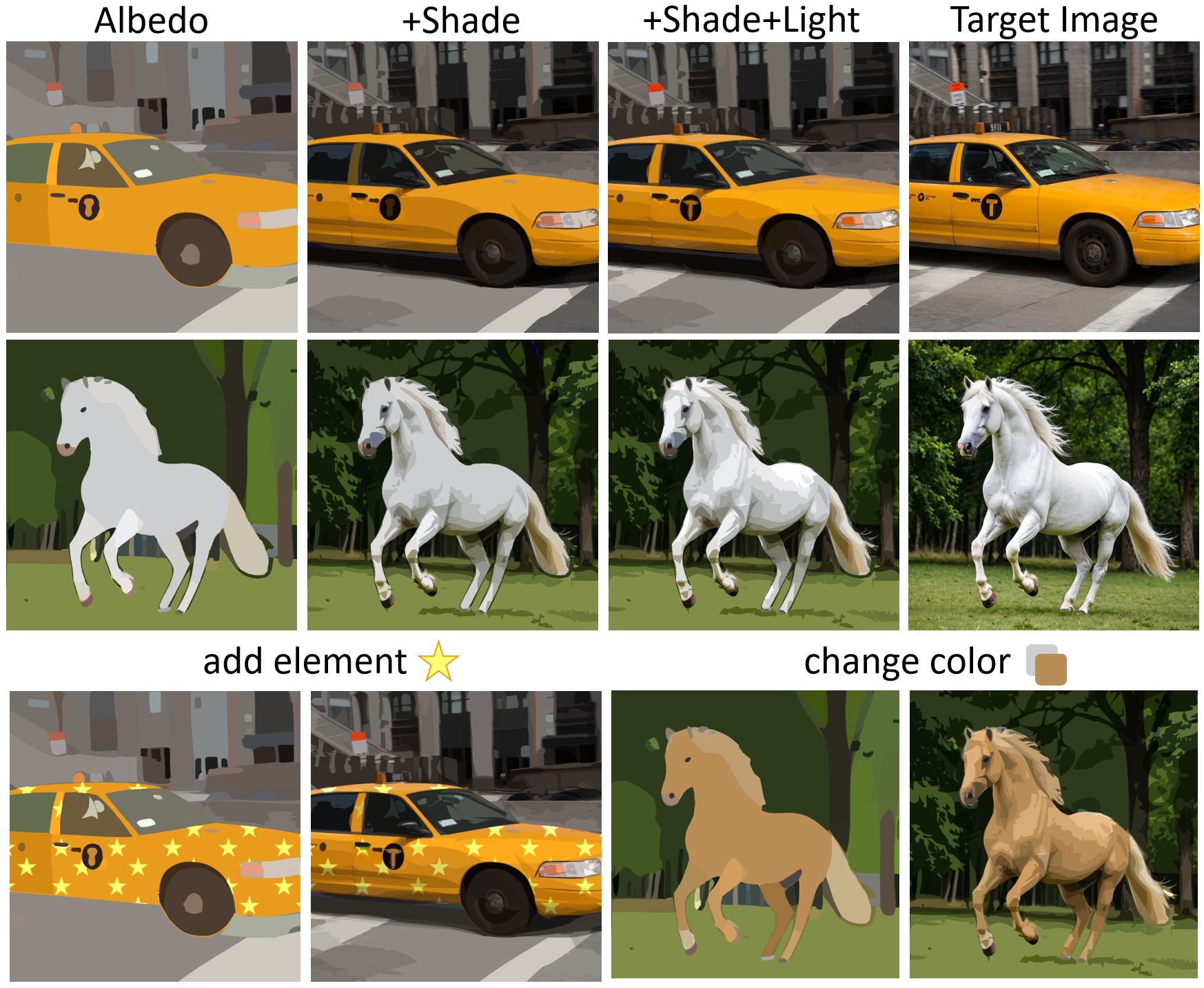}
   \caption{Layer-wise rendering and editing results of our method.  
   The top rows show the progressive composition from albedo, shade, and light layers, revealing how illumination enhances depth and realism.  
   The bottom row demonstrates layer-level editability: modifying only the albedo layer while preserving illumination, enabling flexible edits without disrupting the original light–shade structure.}
   \label{fig:show}
   \vspace{-0.8em}
\end{figure}

\begin{figure}[htbp]
  \centering
   \includegraphics[width=0.85\linewidth]{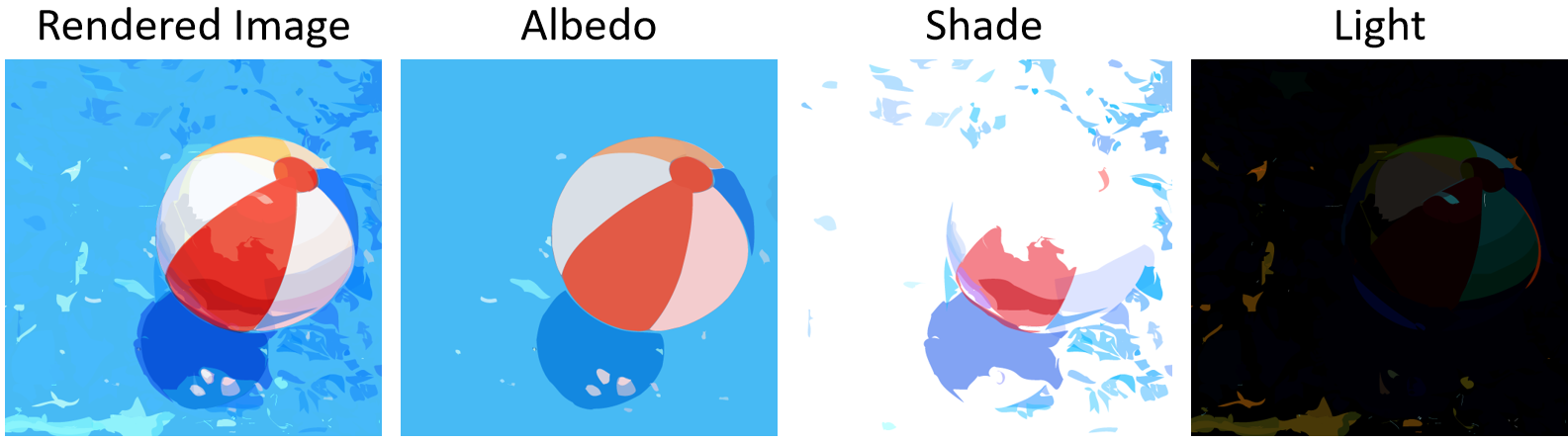}
   \caption{The albedo, shade, and light layers of an image.}
   \label{fig:decompose}
   \vspace{-0.8em}
\end{figure} 

\begin{figure}[htbp]
  \centering
   \includegraphics[width=0.7\linewidth]{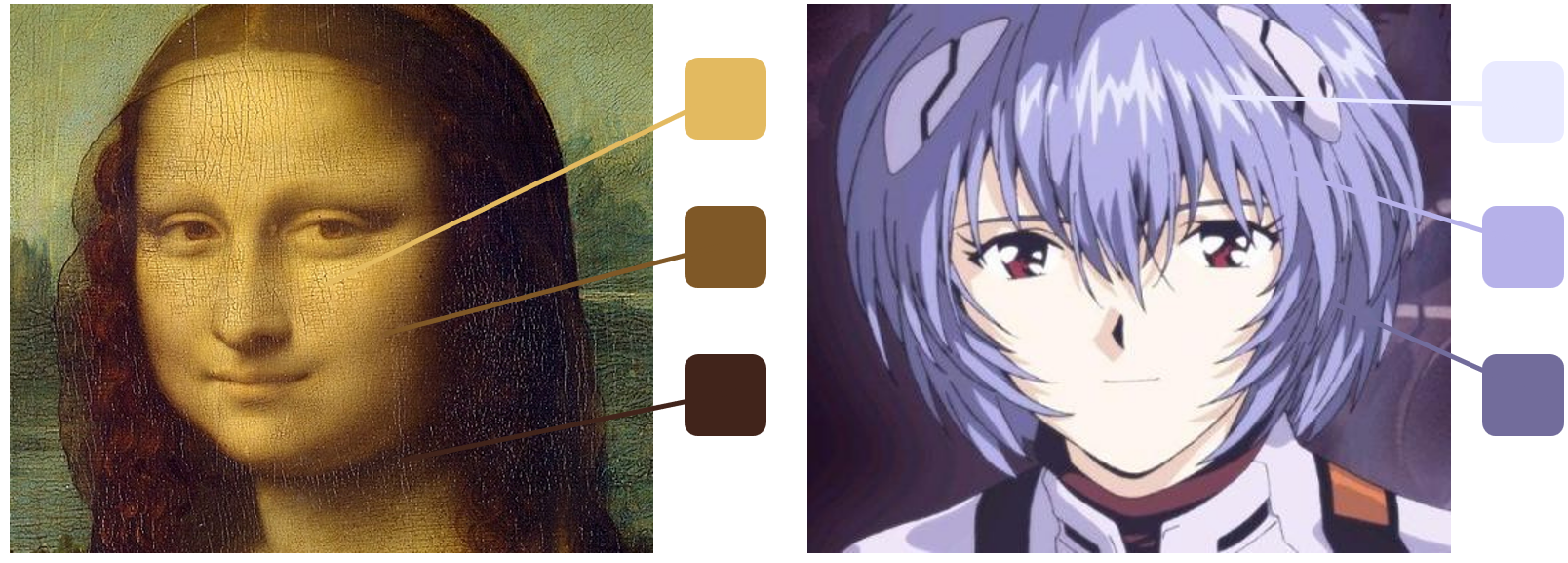}
   \caption{The principle of \textit{Clair-Obscur} in art.  
   Classical painting and modern animation use tone variations within the same semantic regions (e.g. skin, hair) to convey light–shade structure.}
   \label{fig:clair_obscur}
   \vspace{-0.7em}
\end{figure}

In this paper, we focus on real-world image vectorization, aiming at preserving both visual fidelity and semantic conciseness.
Unlike icons or stylized illustrations with uniform tones, real photographs contain rich tonal variation induced by lighting and shadow. Artists traditionally address this complexity through \textit{Clair-Obscur}, an illumination-driven abstraction which convey form and depth by contrasting light and shade, as shown in Figure~\ref{fig:clair_obscur}. 
Drawing inspiration from this principle and recent advances in intrinsic image decomposition~\cite{shah2023join, zhang2024learning, careaga2024colorful}, we decompose images into three semantically meaningful vector layers: albedo, shade and light layers. As shown in Figure~\ref{fig:decompose}, the \textbf{albedo layer} represents intrinsic surface color independent of illumination, the \textbf{shade layer} captures light attenuation, and the \textbf{light layer} models additive highlights and reflections.
This decomposition enables explicit separation of albedo and illumination, producing representations that remain both semantically concise and visually consistent.

Building upon this layered formulation inspired by \textbf{C}lair-\textbf{O}bscur, we introduce \textbf{COVec}, an illumination-aware method for real-world image \textbf{Vec}torization.
COVec decomposes an image into albedo, shade, and light layers within a unified SVG representation.
Specifically, to ensure semantically concise layering, we propose a region-wise semantic binarization strategy to initialize the albedo and illumination layers.
Next, both layers are jointly optimized via differentiable rendering~\cite{li2020differentiable}, enabling consistent reconstruction of color and shading.
Subsequently, the illumination layer is refined individually to capture fine-grained tonal variations and then separated into shade and light components according to highlight localization.
The process ultimately yields a unified SVG representation integrating albedo, shade, and light layers with native SVG blending.

Extensive experiments on various real-world datasets~\cite{karras2019style, hebart2019things, ponomarenko2015image} validate the effectiveness of COVec (Figure~\ref{fig:show}). 
Our method performs well not only on complex real-world images but also on simpler datasets such as emojis.
COVec achieves lower reconstruction and perceptual errors under different path constraints, demonstrating its robustness and visual fidelity.
The semantically concise decomposition enables intuitive and illumination-consistent editing: recoloring the albedo layer naturally adjusts appearance without disrupting illumination structure. 
Compared with existing methods, COVec attains visually coherent results with significantly fewer path edits.
Ablation studies demonstrate the superiority of our proposed region-wise semantic binarization strategy for illumination initialization.

To summarize, our contributions are as follows:
\begin{itemize}
\item We propose COVec, an illumination-aware vectorization method that, for the first time, leverages intrinsic decomposition into image vectorization, enabling consistent and concise reconstruction for real-world images.

\item We design a region-wise semantic binarization strategy to initialize vector layers and demonstrate its effectiveness for illumination modeling.

\item Compared with previous methods, COVec achieves higher visual fidelity and also semantically concise vector representations, resulting in enhanced editability and structural coherence.
\end{itemize}

\section{Related Work}
\label{sec:related work}

\subsection{Image Vectorization}
Image vectorization converts raster images into vector representations composed of geometric primitives such as polygons and Bézier curves.
Early approaches segment regions~\cite{diebel2008bayesian, selinger2003potrace, favreau2017photo2clipart} and approximate their color or contour boundaries with vector shapes,
while later works employ more expressive primitives, such as diffusion curves~\cite{orzan2008diffusion, xie2014hierarchical, zhao2017inverse} and gradient meshes~\cite{sun2007image}, to capture smooth color transitions.

Recent advances in deep learning have brought significant progress to this field.
These methods for generating vector graphics can be broadly categorized into optimization-based~\cite{li2020differentiable, ma2022towards, hirschorn2024optimize, zhou2024segmentation, wang2025layered} and dataset-based approaches~\cite{lopes2019learned, carlier2020deepsvg, rodriguez2025starvector, xing2025empowering, yang2025omnisvg}.
Dataset-based methods rely on large-scale datasets to learn vector generation models.
Early works such as SVG-VAE~\cite{lopes2019learned} and DeepSVG~\cite{carlier2020deepsvg} learn latent representations of simple vector icons, while more recent large-scale models like StarVector~\cite{rodriguez2025starvector} attempt to extend this paradigm using transformer architectures. 
However, due to data scarcity and domain bias, these dataset-based models generalize poorly to real-world images, often producing incomplete or empty SVGs~\cite{rodriguez2025starvector}.
In contrast, optimization-based methods directly fit geometric primitives to an input image without requiring training data. They are highly generalizable and can be applied across diverse visual domains. The differentiable rasterizer DiffVG~\cite{li2020differentiable} has greatly advanced this direction, enabling gradient-based optimization of vector primitives and inspiring subsequent works such as LIVE~\cite{ma2022towards} and LayerVec~\cite{wang2025layered}. 

To further enhance structural organization and editability, some optimization-based works~\cite{ma2022towards, hirschorn2024optimize, wang2025layered, zhou2024segmentation} explore layer decomposition, where an image is represented as multiple vector layers for hierarchical editing. LIVE~\cite{ma2022towards} introduces a bottom-up SVG construction framework that progressively adds and optimizes Bézier paths using differentiable rendering. O\&R~\cite{hirschorn2024optimize} adopts an opposite, top-down strategy, iteratively optimizing and pruning Bézier curves according to their contribution to the overall reconstruction quality.
LayerVec~\cite{wang2025layered} employs a semantically guided, progressive vectorization pipeline. It reconstructs images from coarse to fine guided by a sequence of simplified images.

However, existing layer decomposition techniques mainly operate within RGBA compositing frameworks.
When applied to real-world images, they often produce redundant vector paths that compromise semantic conciseness.
In contrast, our method leverages the rich blending modes supported by the SVG format, such as multiplicative blending to simulate realistic shading effects, to explicitly decouple albedo, shade, and light layers, achieving semantically meaningful and editable vector representations.

\subsection{Intrinsic Image Decomposition}
Intrinsic image decomposition aims to separate an image into its albedo and illumination components, enabling illumination-aware manipulation. 
Classical intrinsic models~\cite{tappen2002recovering, shen2008intrinsic, bell2014intrinsic, li2018cgintrinsics, careaga2023intrinsic} adopt the Lambertian diffuse assumption, expressing an image as a multiplicative combination of albedo and grayscale shading. 
These models are effective for simple scenes but fail to capture colored lighting and specular effects, leading to inaccurate albedo estimation in complex, real-world environments.
To overcome the limitations of grayscale shading, recent works~\cite{das2022pie, meka2021real, lettry2018unsupervised, li2018learning} have developed the model to incorporate RGB shading,  accounting for colored illumination. 
~\citet{careaga2024colorful} adopt an extended image formation model~\cite{shah2023join, zhang2024learning} that decomposes images into albedo, colorful shading, and a residual component for specularities and light sources.
By relaxing the Lambertian and single-color assumptions, it achieves illumination-aware decomposition for in-the-wild photographs.

However, these methods operate purely in the pixel domain, where decomposed components are produced as separate raster images.
These images must be manually combined in professional editing software to achieve a coherent rendering result.
In contrast, we represent albedo, shade, and light as distinct vector layers within a unified SVG output, leveraging SVG’s native blending modes for compositing and editing.

\begin{figure*}[htbp]
  \centering
  \includegraphics[width=0.99\linewidth]{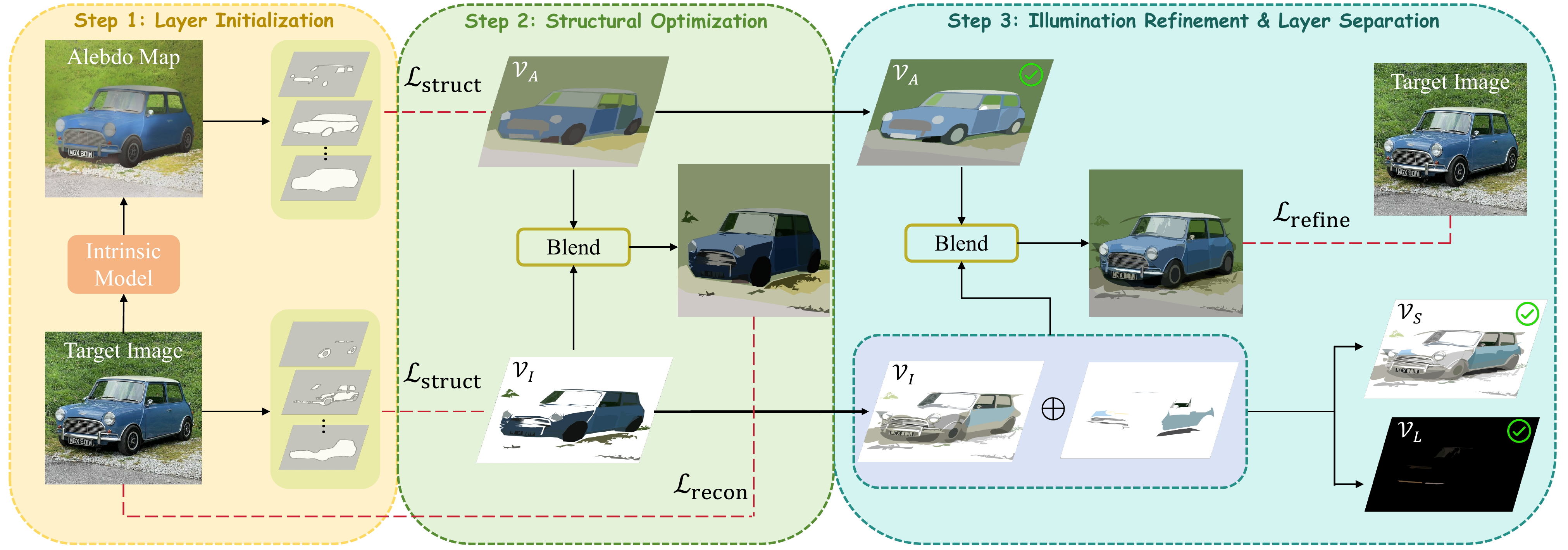}
  \caption{
\textbf{Pipeline of COVec.} We initialize albedo and illumination layers with distinct strategies to obtain their respective masks. 
The first stage jointly optimizes both layers via differentiable rendering, where \(\mathcal{L}_{\text{struct}}\) enforces structural consistency with segemented masks and \(\mathcal{L}_{\text{recon}}\) minimizes the reconstruction error between the blended result (~\cref{eq:render_equation}) and the target image. 
The second stage refines illumination details with \(\mathcal{L}_{\text{refine}}\). 
Finally, the illumination layer is separated into shade and light components to produce a unified SVG output.
  }
  \label{fig:pipeline}
\end{figure*}

\section{Preliminary: Light--Shade Decomposition}

Following the extended image formation model adopted by ~\cite{careaga2024colorful, shah2023join, zhang2024learning}, a pixel image $\mathcal{I}$ can be represented as the combination of intrinsic albedo, shading, and additive lighting:
\begin{equation}
    \mathcal{I} = \mathcal{A} * \mathcal{S} + \mathcal{L},
\end{equation}
where $\mathcal{A}, \mathcal{S}, \mathcal{L}, \mathcal{I} \in [0,1]^{H\times W\times3}$ denote per-pixel RGB albedo, shading, lighting, and observed image, respectively. We adapt this principle to image vectorization.
Specifically, the target SVG representation $\mathcal{V}_{final}$ can be formulated as the composition of three vector layers:
\begin{equation}
\mathcal{V}_{final} = \mathcal{V}_{A} * \mathcal{V}_{S} + \mathcal{V}_{L},
\label{eq: decompose}
\end{equation}
where \(\mathcal{V}_A\), \(\mathcal{V}_S\), and \(\mathcal{V}_L\) denote the albedo, shade, and light layers, which are interpreted as follows: 
\begin{itemize}
\item The albedo layer (\(\mathcal{V}_A\)) represents the intrinsic surface reflectance, capturing colors that remain constant under varying illumination.  
\item The shade layer (\(\mathcal{V}_S\)) models geometric light attenuation caused by visibility and orientation, darkening regions in proportion to shading intensity.
It is composited with the albedo layer using the \texttt{multiply} blending mode.
\item The light layer (\(\mathcal{V}_L\)) accounts for additive illumination effects such as highlights, reflections, and scattered light, composited through the \texttt{plus-lighter} blending mode.  
\end{itemize}

Each decomposition layer \(\mathcal{V}_*\) is represented as a collection of parameterized vector paths:
\begin{equation}
    \mathcal{V}_* = \sum_{n=1}^{N_*} \theta_n^*, \quad \theta_n^* = \{\text{P}_n,\, C_n,\, \tau_n\},
\end{equation}
where a vector path is defined by its geometric shapes \(P_n\), fill color \(C_n\), and opacity \(\tau_n \in [0,1]\).

Practically, directly optimizing the three components in ~\cref{eq: decompose} is challenging due to the strong coupling among variables.
To simplify the optimization, we draw upon the simplified Lambertian intrinsic decomposition~\cite{careaga2024colorful}, where image intensities are modeled as the product of albedo and illumination. We first decompose the objective into two vector layers: the albedo layer and illumination layer, and jointly optimize them:
\begin{equation}
    \mathcal{V}_{final} = \mathcal{V}_A * \mathcal{V}_I,
\label{eq:render_equation}
\end{equation}
where $\mathcal{V}_I$ denotes the illumination layer which combines shading and lighting components. Afterward, we refine it with $\mathcal{V}_A$ fixed and then separate it into shade and light layer.

\section{Method}
\label{sec:method}

\subsection{Overview of Pipeline}




As shown in Figure~\ref{fig:pipeline}, the overall COVec pipeline follows a coarse-to-fine reconstruction process, including initialization and optimization. 
We begin with the initialization of \(\mathcal{V}_A\) and \(\mathcal{V}_{I}\).
To ensure clear layer separation, we adopt distinct initialization strategies for the two decomposition layers.
The albedo layer $\mathcal{V}_A$ is initialized from illumination-invariant albedo masks extracted by a pretrained segmentation model.
In contrast, the illumination layer $\mathcal{V}_I$ is initialized through a region-wise semantic binarization on the target image.

In the first stage of optimization, we jointly optimize \(\mathcal{V}_A\) and \(\mathcal{V}_{I}\) via differentiable rendering~\cite{li2020differentiable}.
The objective contains a structure loss computed with respect to each layer’s mask and a reconstruction loss between the blended composite and the target image.  
Subsequently, in the second stage, \(\mathcal{V}_A\) is fixed, and the illumination layer \(\mathcal{V}_{I}\) is refined by introducing new vector paths and minimizing a refinement loss to capture fine-grained lighting variations.
Finally, we separate the shade and light layers, \(\mathcal{V}_S\) and \(\mathcal{V}_L\), from the illumination layer \(\mathcal{V}_{I}\). 
The resulting three decomposition layers are then composited according to the image formation model described in ~\cref{eq: decompose}, using SVG blending modes to produce a unified and editable vector output.

\subsection{Layer Initialization}

A good initialization is crucial for efficient optimization~\cite{ma2022towards, hirschorn2024optimize}, as it provides a reliable starting direction and significantly reduces optimization time. 
To enable vector-based optimization, we initialize the albedo and illumination layers as explicit regions.

\textbf{Albedo Masks.}
The albedo layer represents the intrinsic reflectance of surfaces. 
We first obtain the albedo map from the target image using a standard intrinsic decomposition model~\cite{careaga2024colorful}.
However, directly using the raw albedo intensity does not provide clear structural regions. 
To extract semantic structural regions that are unaffected by illumination variations, we employ a pretrained segmentation model, SAM~\cite{kirillov2023segment}, on the albedo map to generate albedo masks. 
These masks define the vector regions for \(\mathcal{V}_A\).

\begin{figure}[htbp]
  \centering
   \includegraphics[width=0.9\linewidth]{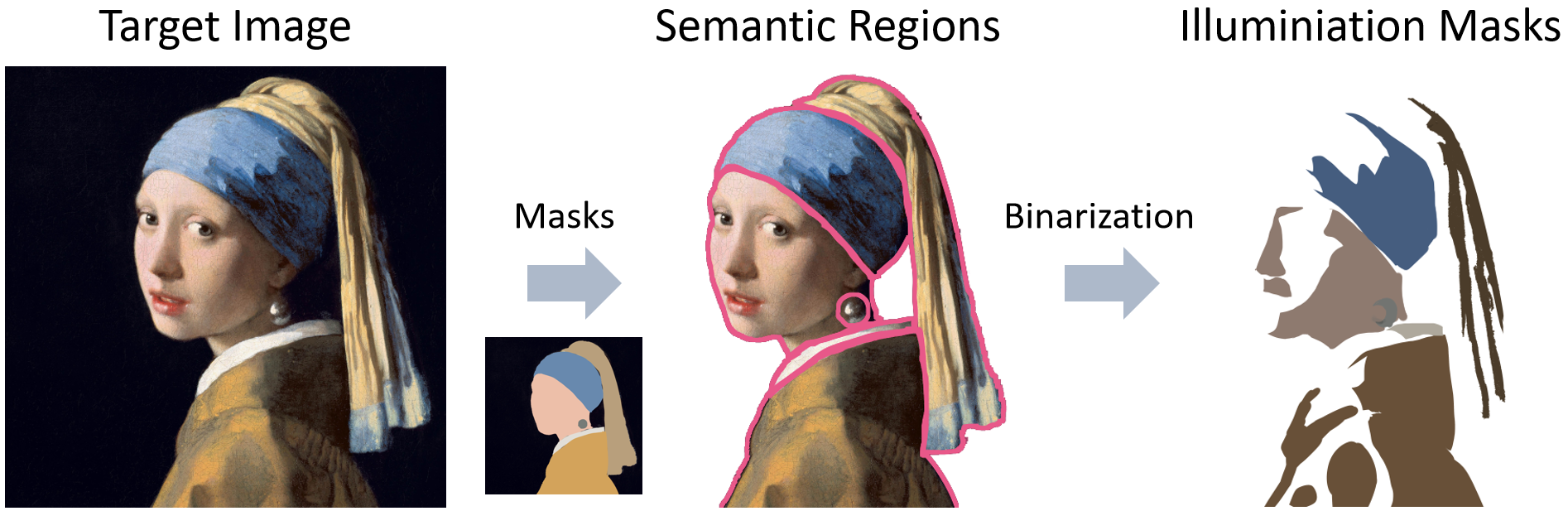}
   \caption{Illustration of the region-wise semantic binarization process. The target image is binarized within each semantic mask, yielding illumination masks aligned with object boundaries.}
   \label{fig:binary}
\end{figure}

\textbf{Illumination Masks.} 
The illumination layer provides an estimate of the illumination components, including shading and lighting.
At the initialization stage, the illumination layer is approximated by extracting shadow regions from the target image. 
We observe that the intensity of shadows varies across different regions of an image.
Therefore, we propose a \textbf{region-wise semantic binarization strategy} to adaptively capture such spatial variations.
As shown in Figure~\ref{fig:binary}, instead of thresholding the entire image uniformly, we perform binarization independently within each semantic region of the target image. 
Pixels outside these semantic regions are ignored, ensuring that binarization operates only within meaningful object regions.

Let $\{ m_1^A, m_2^A, \dots, m_n^A \} \subset \mathbb{R}^{w \times h}$ denote the set of albedo masks corresponding to different semantic regions in the target image. 
For each albedo mask $m_i^A$, a region-wise threshold is computed as the mean pixel intensity within that region:
\begin{equation}
    T_i = \frac{1}{|m_i^A|} \sum_{p \in m_i^A} \mathcal{I}(p),
\end{equation}
where $\mathcal{I}(p)$ denotes the pixel intensity of the target image $\mathcal{I}$ at pixel $p$. 
Pixels of the target image that fall within $m_i^A$ and satisfy $\mathcal{I}(p) \le T_i$ are classified as shadow pixels, forming illumination masks. 
These masks define the vector regions for \(\mathcal{V}_I\).

\textbf{Organization of Segmented Masks.}
After obtaining the masks of the two layers, we organize them into hierarchical mask groups following a back-to-front ordering strategy from LayerVec~\cite{wang2025layered}. 
Specifically, mask organization is performed independently within each layer.
Larger masks are placed in bottom groups, while smaller ones are placed on top to avoid overlap within a single group. 
Each mask boundary is extracted and simplified using the Douglas–Peucker algorithm~\cite{douglas1973algorithms} to reduce redundant points while preserving structural fidelity. 
For each mask, the corresponding vector shape is represented as a closed cubic Bézier contour, with control points sampled from the simplified boundary.

After the initialization and organization of segmented masks, both the albedo and illumination layers consist of multiple vector path groups organized hierarchically.

\subsection{Structural Optimization}
\label{subsec:Struct_illum}

To effectively optimize the two-layer vector decomposition, we adopt a progressive strategy that first stabilizes the geometric structure of each layer and then refines their joint blended result. 

\textbf{Structure Loss.} 
We adopt a layer-wise structure loss to ensure geometric consistency between the vectorized representation and its mask group. 
For each path group \(\mathcal{I}^{\text{group}}_{j}\) and its mask group \(\mathcal{I}^{\text{mask}}_{j}\), the structure loss is formulated as:
\begin{small}
\begin{equation}
    \mathcal{L}_{\text{struct}} = 
    \sum_{j=1}^{N} (\left\| \mathcal{I}^{\text{mask}}_{j} - \mathcal{I}^{\text{group}}_{j} \right\|_2^2
    + \lambda\sum_{p \in I_{\text{group},j}'} \text{ReLU}\left(\delta - \alpha(p)\right)),
\end{equation}
\end{small}
where \(N\) is the number of path groups, vectors are rendered with the same semi-transparent gray in \(I_{\text{group},j}'\), \(\alpha(p)\) is the pixel transparency of pixel \(p\), \(\delta\) is the threshold controlling overlap tolerance, and $\lambda$ is a weighting factor set to $1\times10^{-8}$. 
The first term enforces shape alignment between the vector paths and reference masks with pixel-wise MSE loss, while the second term discourages excessive self-overlap within the same group. 
This structure loss guides each path group toward compact and non-redundant geometric representations.

\textbf{Reconstruction Loss.}
To encourage overall visual consistency, we define a shared reconstruction loss, which measures the difference between the rendered composite and the target image:
\begin{equation}
    \mathcal{L}_{\text{recon}} =
    \left\| \mathcal{I} - \mathcal{R}(\mathcal{V}_{A}) * \mathcal{R}(\mathcal{V}_{I}) \right\|_2^2,
\label{eq:recon_loss}
\end{equation}
where \(\mathcal{R}(\cdot)\) denotes the differentiable rendering operator~\cite{li2020differentiable}.

\textbf{Optimization Schedule.}
We optimize the albedo layer $\mathcal{V}_{A}$ and the illumination layer $\mathcal{V}_{I}$ using two independent optimizers. 
Optimization follows a two-stage schedule. 
During the warm-up stage, each layer is supervised by its own structure loss,
\begin{equation}
    \mathcal{L}_{A} = \mathcal{L}_{\text{struct}}^{A}, \qquad
    \mathcal{L}_{I} = \mathcal{L}_{\text{struct}}^{I},
\end{equation}
ensuring stable alignment with their initialization masks, guided by the albedo masks and illumination masks respectively.
After the first 50 epochs, we switch to the shared reconstruction objective for the remaining 50 epochs,
\begin{equation}
    \mathcal{L}_{A} = \mathcal{L}_{I} = \mathcal{L}_{\text{recon}},
\end{equation}
while still updating the parameters of $\mathcal{V}_{A}$ and $\mathcal{V}_{I}$ with their own optimizers.
This schedule first stabilizes layer geometry, then jointly refines all layers to reproduce the target image.

\subsection{Illumination Refinement \& Layer Separation}

\textbf{Illumination Refinement.}
With both the albedo layer and the existing illumination base fixed, new vector paths are progressively appended to the illumination layer to capture fine-grained details.  
These newly introduced paths collectively form an enhanced illumination layer.  
Following the path addition strategy of LIVE~\cite{ma2022towards}, new paths are inserted into regions with high reconstruction error.  
During this refinement, only the parameters of the newly added paths are optimized, while previously optimized paths remain fixed.  
The optimization minimizes the reconstruction loss between the rendered composite and the target image:
\begin{equation}
    \mathcal{L}_{\text{refine}} =
    \left\| \mathcal{I} - \mathcal{R}(\mathcal{V}_{A}) * \mathcal{R}(\mathcal{V}_{I}) \right\|_2^2,
\end{equation}
which is the same as \cref{eq:recon_loss}, only with the parameters of $\mathcal{V}_{A}$ freezed.
To maintain a compact representation, we follow LayerVec~\cite{wang2025layered} and periodically perform vector cleanup operations, such as merging and removing redundant paths.

\textbf{Layer Separation.}
After illumination refinement, the illumination layer $\mathcal{V}_{I}$ is separated into the shade layer $\mathcal{V}_{S}$ and the light layer $\mathcal{V}_{L}$. 
The separation is performed by examining each vector path's fill color intensity: 
paths whose color values remain within the normalized range $[0,1]$ are assigned to the shade layer $\mathcal{V}_S$, whereas paths whose values exceed this range are identified as highlight-contributing regions and thus grouped into the light layer $\mathcal{V}_L$.

During this process, the control points of all paths remain unchanged. 
Paths belonging to $\mathcal{V}_{S}$ directly inherit both the shapes and colors from $\mathcal{V}_{I}$, while paths in $\mathcal{V}_{L}$ retain only their shapes. 
Because the light layer uses additive blending, their colors cannot be taken from $\mathcal{V}_{I}$; instead, they are recomputed from the residual illumination. 
Specifically, we compute a residual map $ \mathcal{I} - \mathcal{R}(\mathcal{V}_{A}) * \mathcal{R}(\mathcal{V}_{S})$, and the final color of each light-layer path is obtained by averaging the pixel values of the residual map within the region covered by that path.

\section{Experiments}
\label{sec:exp}

\subsection{Experimental Setup}

\textbf{Implementation Details.}
We implement COVec in PyTorch~\cite{paszke2019pytorch} and optimize it using the Adam optimizer~\cite{kingma2014adam}, 
with a learning rate of \(1.0\) for control point optimization and \(0.01\) for color parameter optimization. 
All experiments are conducted on four NVIDIA L40S GPUs. 
Additional implementation details are provided in the supplementary material.

\textbf{Dataset.}
For evaluation, we collect three datasets, each containing 100 randomly selected images. 
The first two focus on complex real-world images, while the third targets simple images.

\textit{Face Dataset.}
This dataset is derived from the FFHQ dataset~\cite{karras2019style}, which contains high-quality facial photographs under diverse ethnicities, genders, and ages. 

\textit{Scene Dataset.}
We collect images from the Things dataset~\cite{hebart2019things} and TID2013 dataset~\cite{ponomarenko2015image}. 
This dataset includes a wide variety of objects, animals, and natural scenes.

\textit{Emoji Dataset.}
Following O\&R~\cite{hirschorn2024optimize}, we build an emoji dataset from the Noto Emoji dataset~\cite{googlefonts_noto_emoji}, which consists of simple icons with clean color boundaries. 

\textbf{Model Variant for simple images.}
Since the emoji dataset contains visually simple images without complex illumination variation, 
we use a simplified version of COVec that optimizes only a single albedo layer without introducing additional light or shade layers. 
This degenerated configuration reduces COVec to a LayerVec-like pipeline~\cite{wang2025layered},
while still incorporating our initialization strategy. The binary masks obtained from semantic segmentation are merged into the albedo initialization process.

\textbf{Compared Methods.}
We compare COVec with four representative vectorization methods: 
DiffVG~\cite{li2020differentiable}, LIVE~\cite{ma2022towards}, 
O\&R~\cite{hirschorn2024optimize}, and LayerVec~\cite{wang2025layered}. 
Among them, LayerVec represents the current state-of-the-art approach for layered image vectorization. 
For a fair comparison, we adopt Bézier curves as the unified vector primitive and use the same learning rate settings as in COVec across all baselines.

\subsection{Main Results}
\textbf{Visual Fidelity.}
We evaluate COVec against the compared methods in terms of visual fidelity, measuring the differences between target images and rendered images.

\begin{figure}[htbp]
  \centering
   \includegraphics[width=0.98\linewidth]{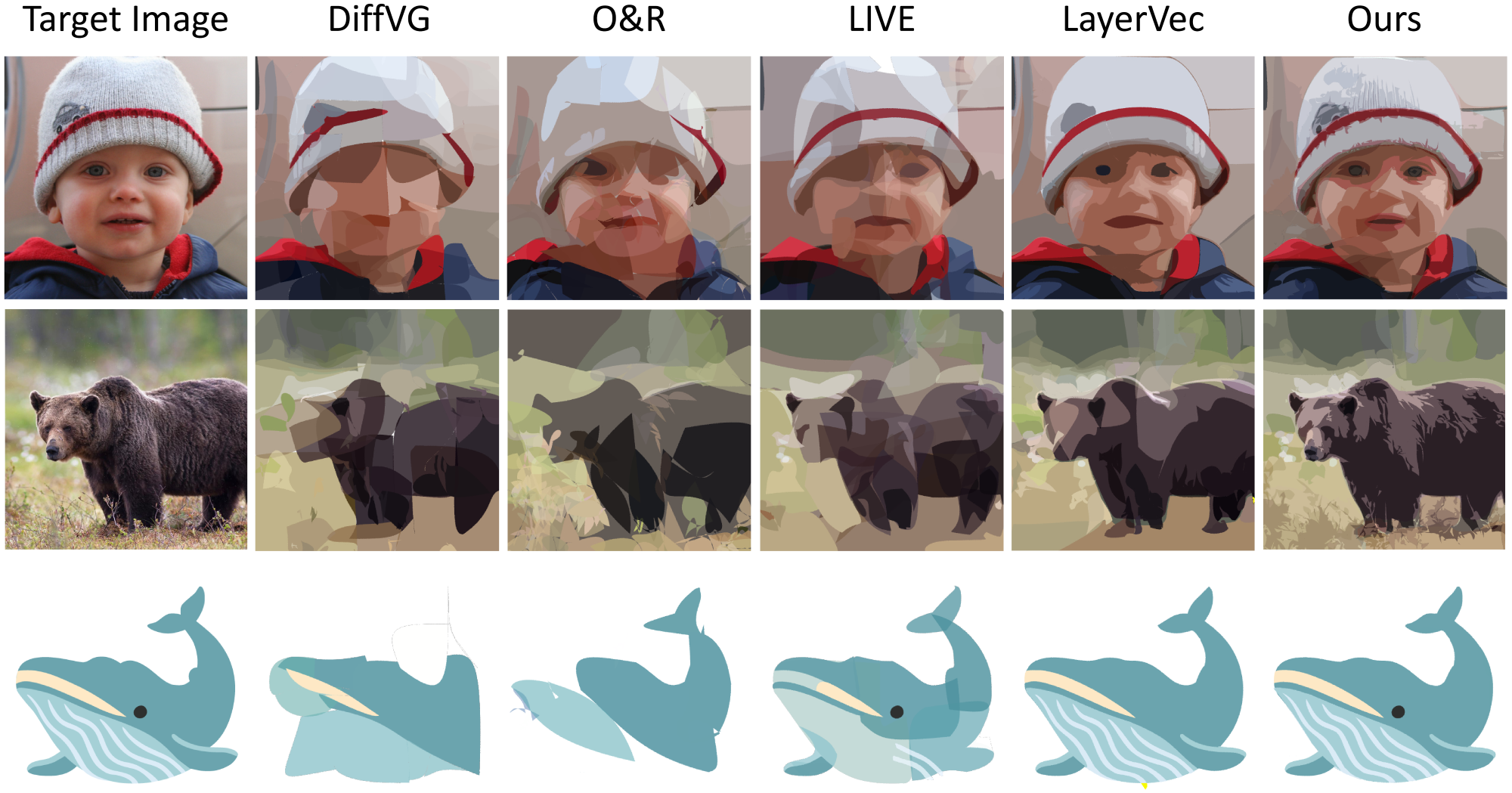}
   \caption{Qualitative reconstruction comparison across datasets. All examples in the Face and Scene datasets are vectorized using 64 paths, whereas Emoji examples are vectorized with 16 paths.
   COVec achieves higher visual fidelity and compact vector structures compared with existing methods.}
   \label{fig:onecol}
\end{figure}

\textit{Qualitative Comparison.}
Figure~\ref{fig:onecol} presents visual comparisons of reconstructed vector images under the same path budgets. 
In complex real-world scenes (top and middle rows), COVec is able to reconstruct complete semantic regions of the target image, such as the child’s eyes and mouth. 
Moreover, COVec preserves fine-grained illumination details that other methods fail to capture. 
While existing approaches rely on redundant paths to approximate these details, COVec uses compact and well-aligned shapes that accurately represent challenging details, such as the fur of the bear.
In the simple emoji domain (bottom row), COVec produces accurate shape contours comparable to LayerVec, validating its adaptability to simplified scenarios.

\begin{figure}[htbp]
  \centering
   \includegraphics[width=0.99\linewidth]{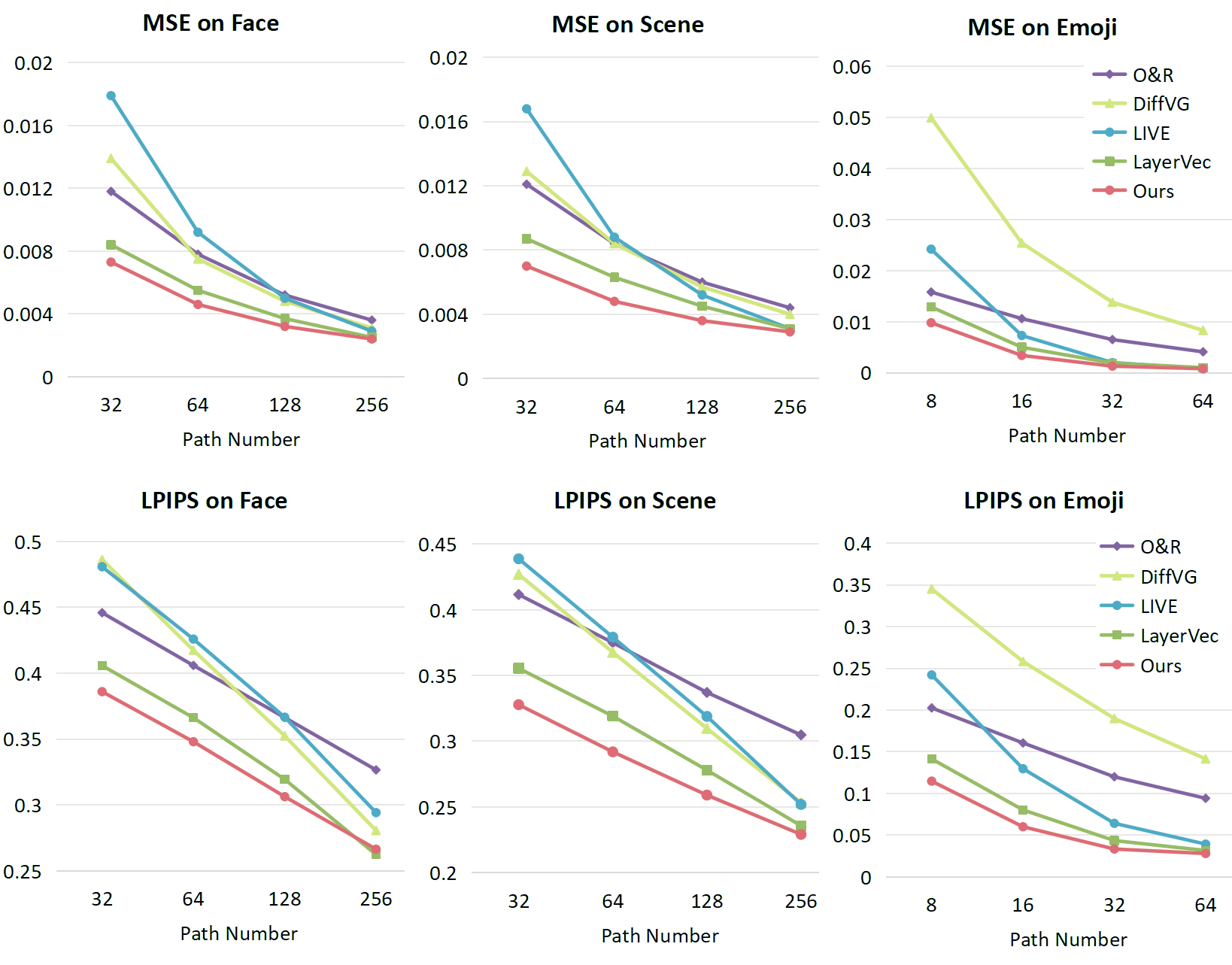}
   \caption{Quantitative comparison in terms of MSE and LPIPS~\cite{zhang2018unreasonable} under different numbers of paths. }
   \label{fig:mse}
\end{figure}

\textit{Quantitative Comparison.}
Figure~\ref{fig:mse} reports both Mean Squared Error (MSE) and Learned Perceptual Image Patch Similarity (LPIPS)~\cite{zhang2018unreasonable} curves under increasing path numbers across all datasets. 
In most cases, COVec achieves the lowest reconstruction error and perceptual distance, consistently outperforming other methods under different numbers of paths.
From complex real-world scenes to simple emoji graphics, COVec attains higher fidelity than LayerVec and O\&R, achieving comparable or better results with substantially fewer paths.


\begin{figure}[htbp]
  \centering
  \includegraphics[width=0.9\linewidth]{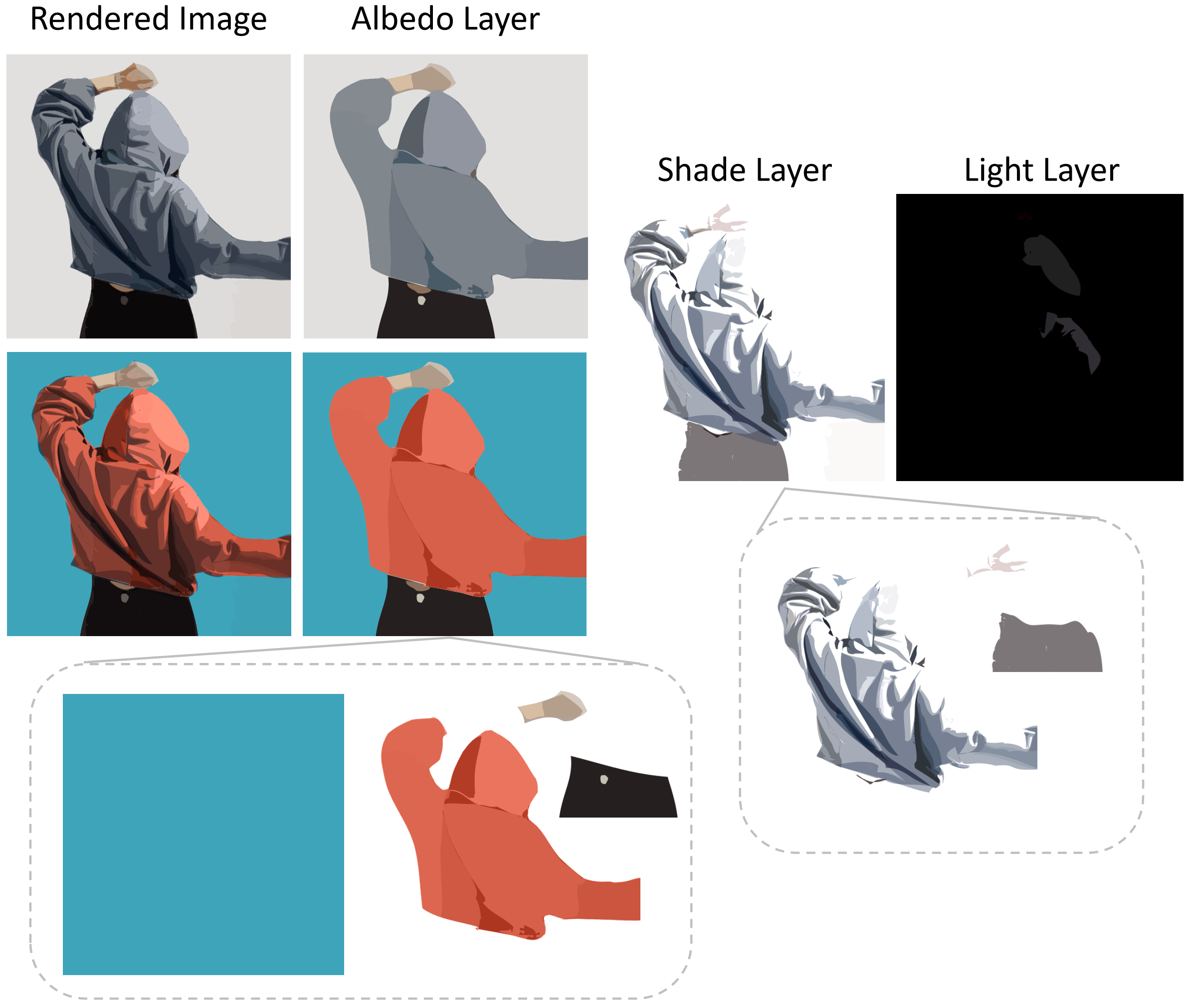}
  \caption{
  Layer decomposition and editing.
  COVec decomposes an image into albedo, shade, and light layers.
  We recolor the albedo layer while keeping the shade and light layers fixed.
  It preserves the illumination structure, while the blending mechanism naturally adapts illumination colors to the new albedo tone.
  }
  \label{fig:vis_decompose}
\end{figure}

\textbf{Layer-wise Representation.}
To demonstrate the semantic conciseness and interpretability of our layered representation, we visualize the decomposed albedo, shade, and light layers in Figure~\ref{fig:vis_decompose}.
COVec explicitly separates intrinsic color and illumination, allowing intuitive editing at the layer level.
By recoloring only the albedo layer while keeping the shade and light layers fixed, the overall illumination effects remain consistent.
It enables flexible appearance modification without breaking illumination coherence.
Moreover, thanks to the hierarchical structure of vector paths, the layers can be further decomposed into compact, part-level representations\textemdash each corresponding to meaningful semantic components.
This concise organization enables efficient localized edits, such as recoloring or relighting specific parts.

\begin{figure}[htbp]
  \centering
  \includegraphics[width=0.95\linewidth]{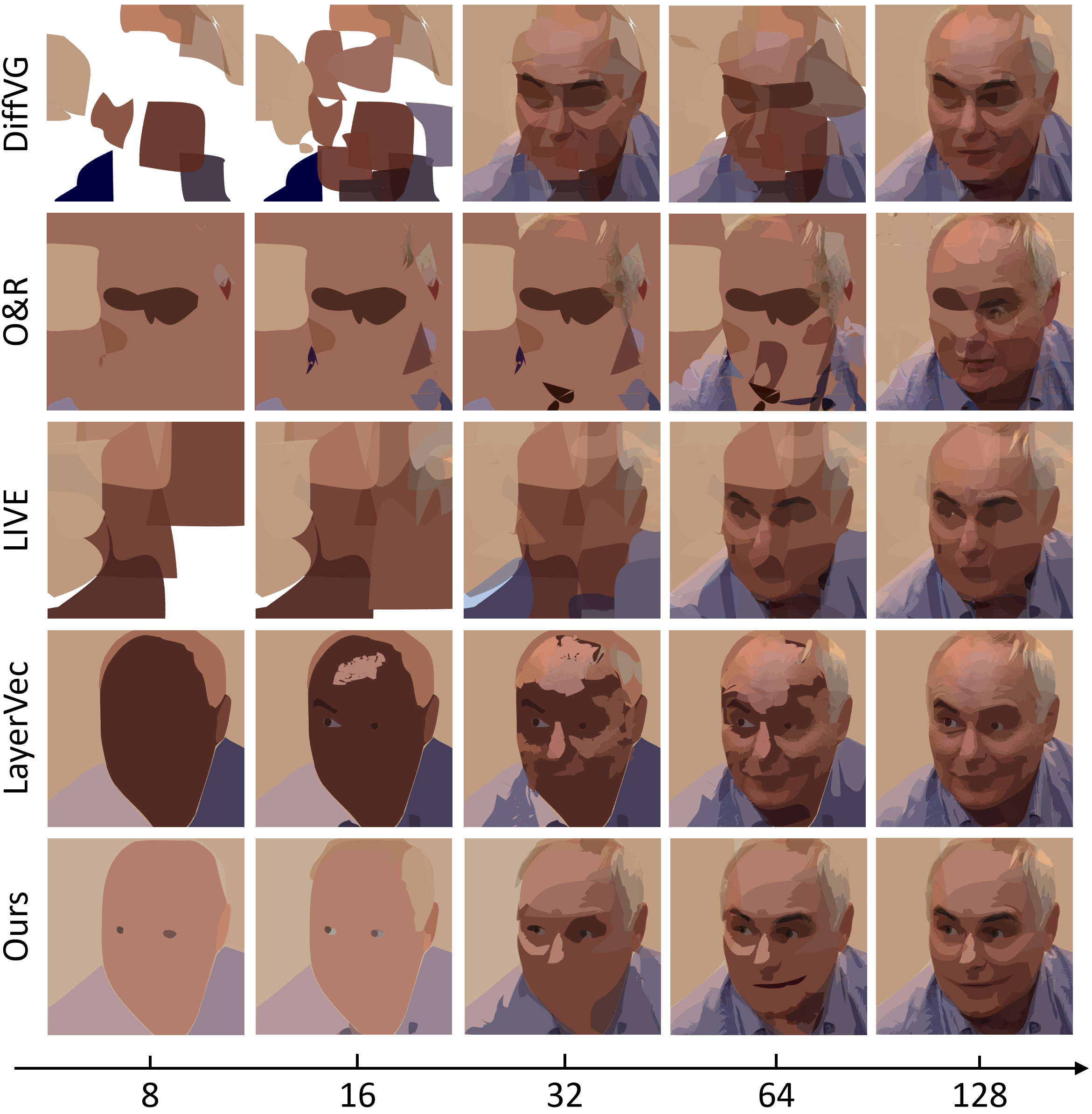}
  \caption{
  Comparison of layer-wise representation.
  Each column shows the result with an increasing number of paths, revealing how the layered representation is progressively organized.
  }
  \label{fig:layer_compare}
  \vspace{-0.3em}
\end{figure}

A qualitative comparison in Figure~\ref{fig:layer_compare} further shows that COVec yields cleaner structural organization across layers compared to existing methods.  
Previous methods rely on fragmented paths to approximate the image, where illumination is implicitly embedded within color regions.
In contrast, our method constructs illumination in a progressive and disentangled manner: it first establishes intrinsic albedo to capture material identity, and then gradually adds paths for the shade and light layers.
This process mirrors the Clair-Obscur painting principle, resulting in semantically concise and illumination-aware vector representations.

\begin{figure}[htbp]
  \centering
  \includegraphics[width=0.93\linewidth]{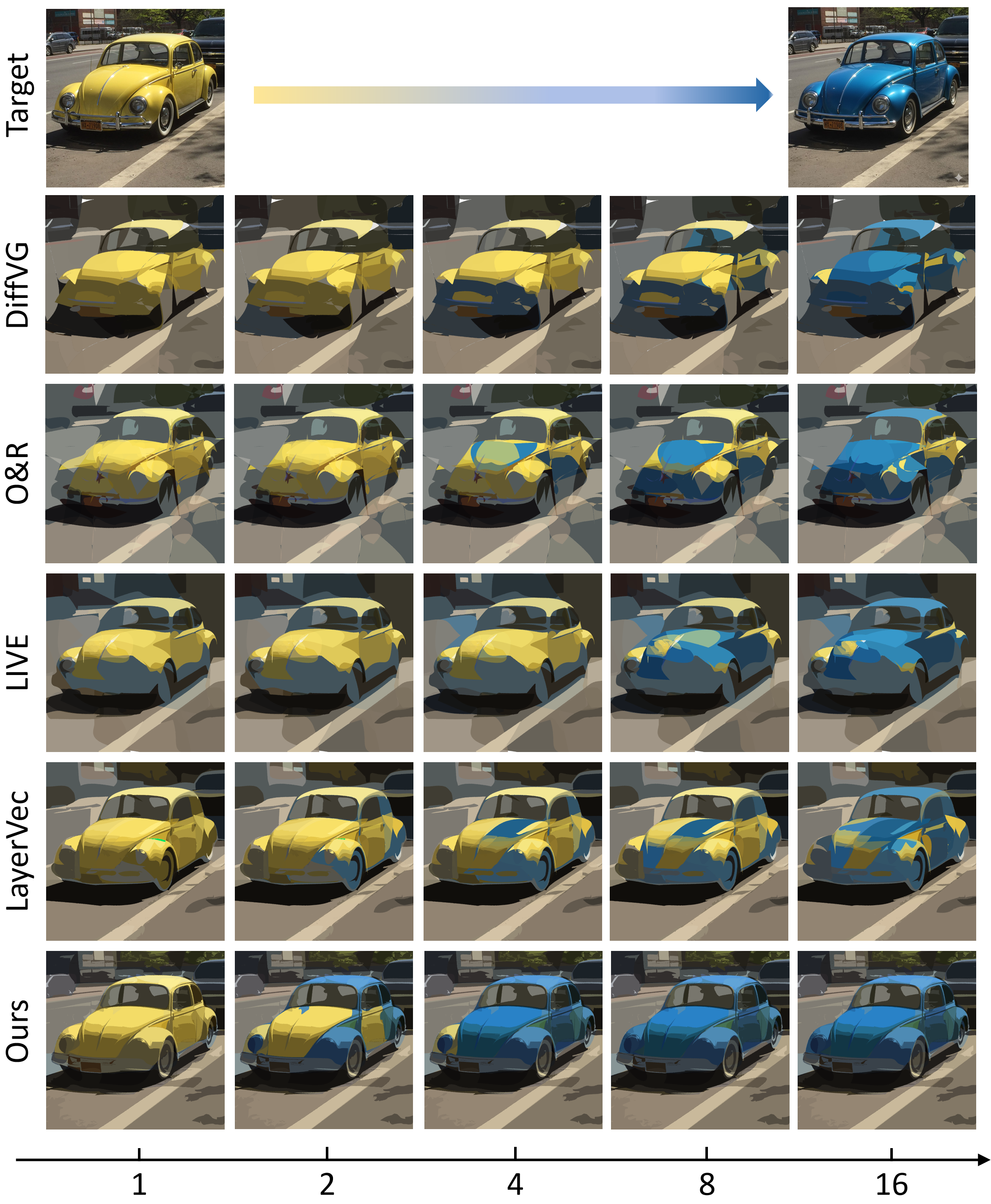}
  \caption{
  Color-editing comparison with controlled path modification.
  Each column corresponds to an increasing number of edited paths (1, 2, 4, 8, 16). The edited paths are selected as the top-k ones with the largest overlaps with the difference regions, and recolored to match the target image.
  }
  \label{fig:edit_compare}
  \vspace{-0.2em}
\end{figure}

\textbf{Image Editing}
To assess the editability of the resulting SVG generated by different vectorization methods, we conduct a controlled color-editing experiment as shown in Figure~\ref{fig:edit_compare}.
For each method, we progressively edit a small number of vector paths according to the color changes suggested by a reference image generated by Gemini 2.5~\cite{comanici2025gemini}.
This process detects color-difference regions between the original and target images and adjusts the corresponding vector paths accordingly.

As the number of edited paths increases, the SVG generated by COVec exhibits the most effective and natural transition towards the target image.
In contrast, other methods produce redundant paths and do not decouple intrinsic color from illumination, making it difficult to achieve the target appearance by editing only a small subset of paths. 
Benefiting from our layered decomposition, COVec can edit intrinsic colors without disrupting the illumination structure, allowing precise edits with only a few path modifications.

\subsection{Ablation Study}
\begin{figure}[htbp]
  \centering
  \includegraphics[width=0.8\linewidth]{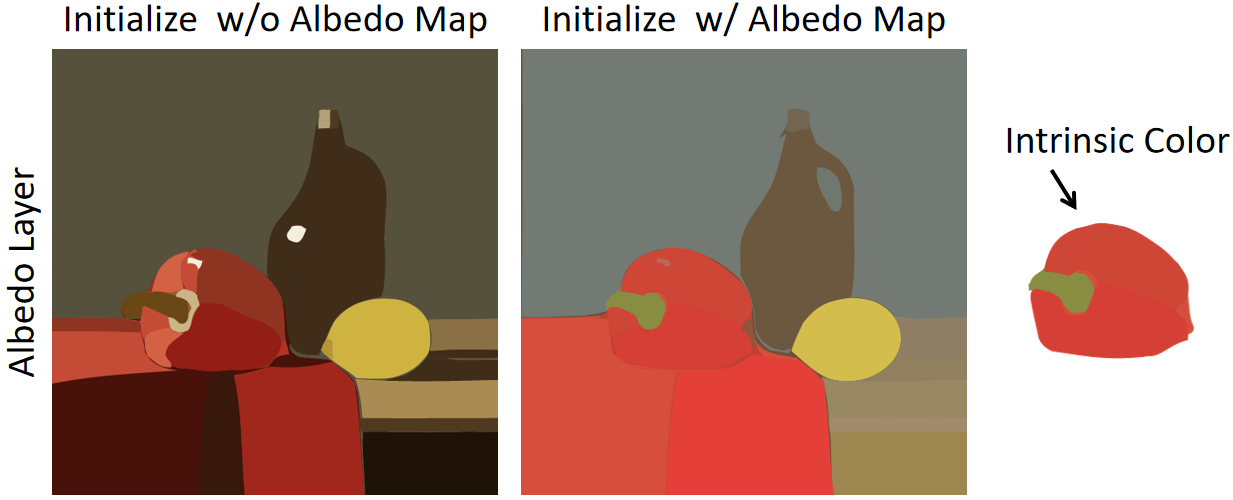}
  \caption{
  Effect of albedo-based initialization.
  Using the albedo map as initialization enables illumination-invariant vector segmentation and clearer intrinsic color separation.
  }
  \label{fig:ablation_albedo}
\end{figure}

\textbf{Albedo-based Initialization.}
A good initialization is essential for achieving stable layer separation and optimization.  
As shown in Figure~\ref{fig:ablation_albedo}, using the intrinsic albedo map to initialize the albedo layer provides a clear structural prior, allowing the vector paths to capture material colors that are invariant to illumination.  
Without the albedo map, the layer optimization tends to entangle intrinsic surface colors with illumination effects.  

\begin{figure}[htbp]
  \centering
  \includegraphics[width=0.99\linewidth]{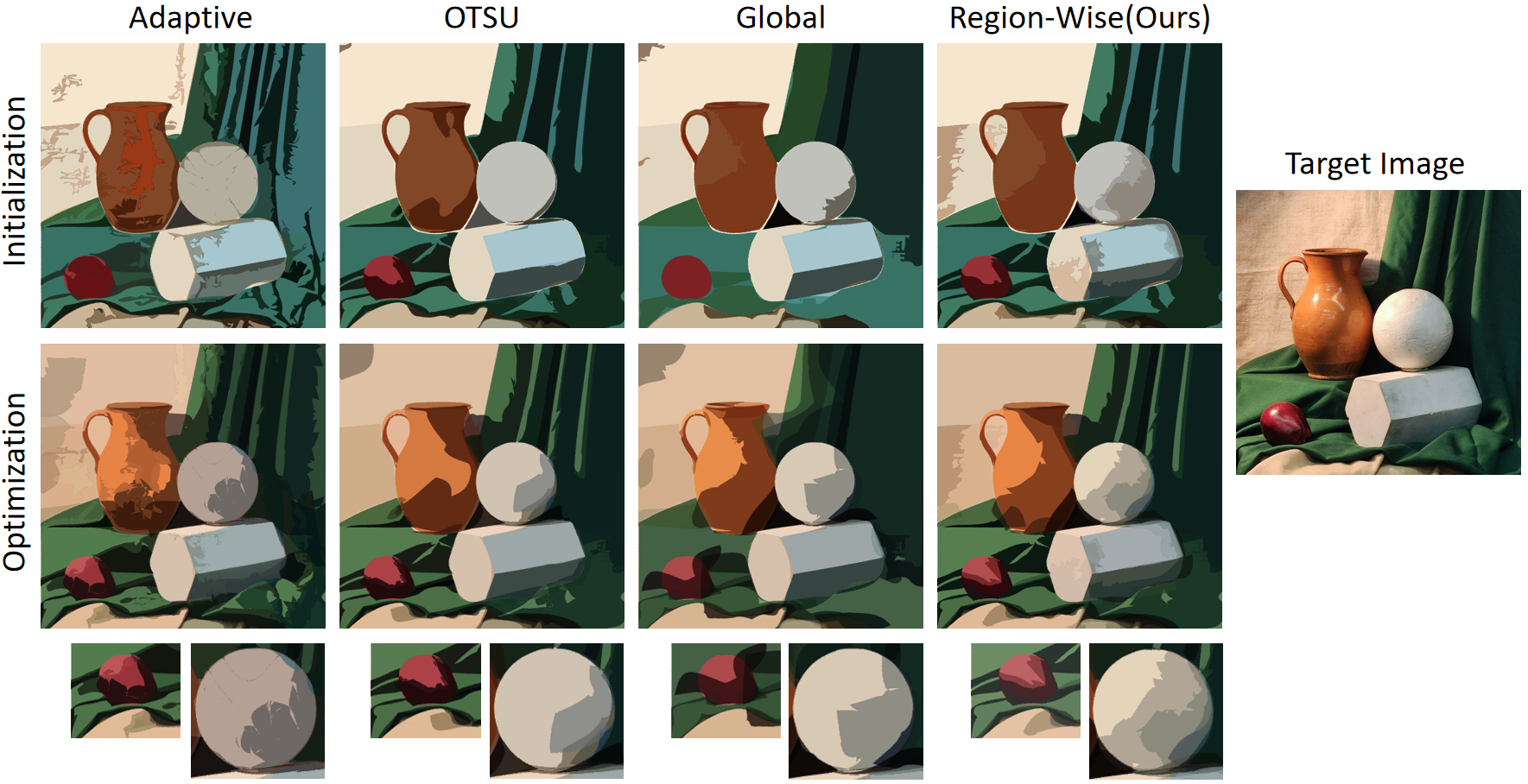}
  \caption{
  Comparison of binarization strategies for illumination initialization.
  Traditional binarization methods, including global thresholding, OTSU, and adaptive thresholding, perform pixel-level thresholding based on global or local intensity distributions.
  }
  \label{fig:ablation_shade}
\end{figure}

\textbf{Region-wise Semantic Binarization.}
To obtain a structurally meaningful illumination layer, we compare four binarization strategies\textemdash adaptive~\cite{sauvola2000adaptive}, OTSU~\cite{otsu1975threshold}, global thresholding, and our proposed region-wise semantic binarization strategy. As shown in Figure~\ref{fig:ablation_shade}, 
traditional pixel-based thresholding fails to align shading regions with object semantics, producing fragmented or inconsistent shadow areas.  
In contrast, our region-wise strategy performs local thresholding within each semantic mask, ensuring that shadow boundaries conform to the geometry of individual objects.  
This alignment results in cleaner initialization and more accurate structural optimization.

\begin{figure}[htbp]
  \centering
  \includegraphics[width=0.8\linewidth]{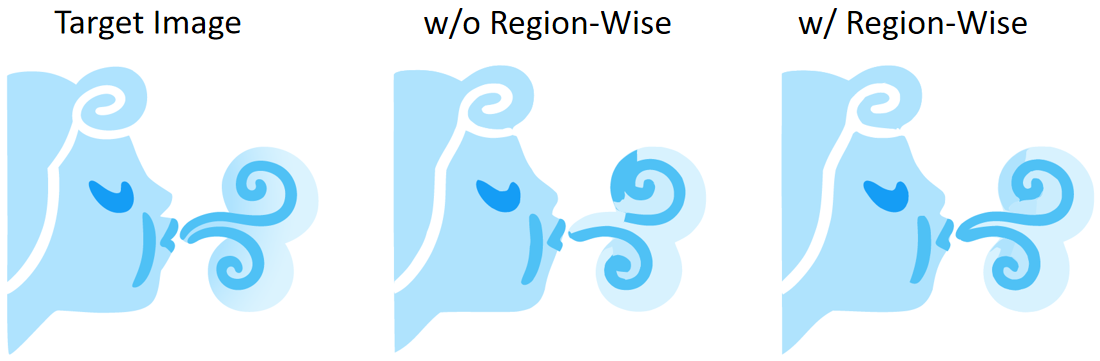}
  \caption{
Effects of region-wise semantic binarization on emoji vectorization results under the same path budget.
  }
  \label{fig:simple}
\end{figure}

We also incorporate region-wise semantic binarization into the initialization of the simplified setting for emoji-style images.
As shown in Figure~\ref{fig:simple}, this strategy preserves finer details and produces cleaner shapes.

\begin{figure}[htbp]
  \centering
  \includegraphics[width=0.9\linewidth]{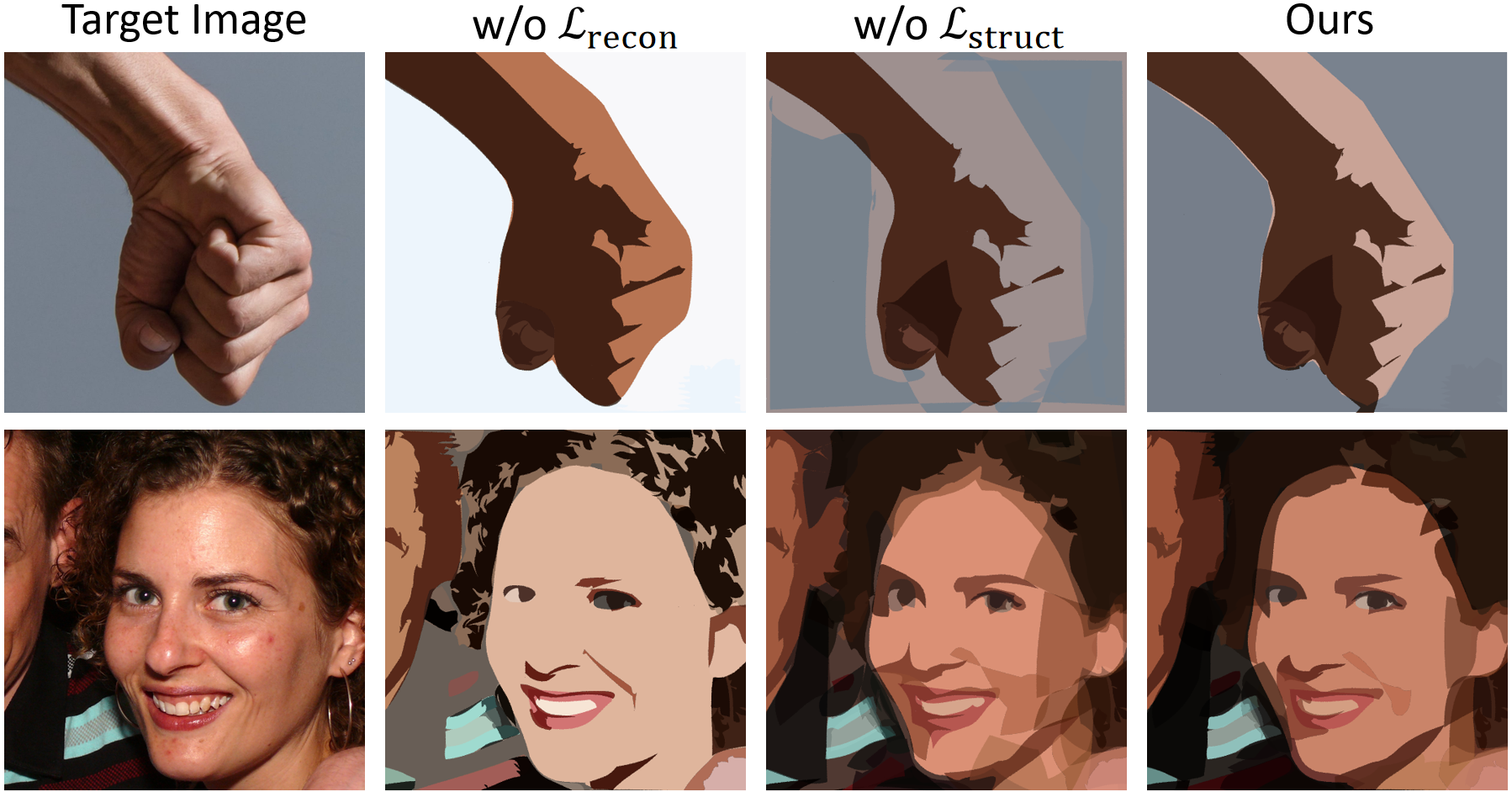}
  \caption{
Ablation of the structure and reconstruction losses for structural optimization. 
  }
  \label{fig:ablation_opt}
\end{figure}

\textbf{Structural Optimization.}
Figure~\ref{fig:ablation_opt} evaluates the effect of removing each component of our optimization objective in structural optimization. 
Without the reconstruction loss $\mathcal{L}_{\text{recon}}$, the model fails to recover accurate colors, yielding visually inconsistent results despite relatively stable geometry.
On the other hand, removing the structure loss $\mathcal{L}_{\text{struct}}$ leads to irregular and unstable vector boundaries, causing fragmented regions and degraded semantic coherence, although the overall color tone is preserved. 
Our full objective contains both losses, yielding smooth and well-organized vector geometry together with faithful color that best matches the target image. This joint formulation provides a stable foundation for subsequent refinement.

\section{Conclusion}
\label{sec:conclusion}

We propose COVec, an illumination-aware method for real-world image vectorization that integrates the artistic principle of Clair-Obscur into the vector domain. 
COVec bridges intrinsic image decomposition with geometric vectorization, explicitly decoupling albedo, shade, and light into distinct layers within a unified vector representation.  
We propose a region-wise semantic binarization strategy that serves as an effective initialization for illumination modeling. 
Extensive experiments demonstrate that our method achieves superior visual fidelity, more semantically concise layer organization, and stronger editability compared with existing approaches.

\section*{Acknowledgements}
This work is supported by the project of National Natural Science Foundation of China (No. 62376012), which is also a research achievement of State Key Laboratory of Multimedia Information Processing, National Engineering Research Center of New Electronic Publishing Technologies and Key Laboratory of Science, Technology and Standard in Press Industry (Key Laboratory of Intelligent Press Media Technology).

{
    \small
    \bibliographystyle{ieeenat_fullname}
    \bibliography{main}
}

\clearpage
\appendix

\section{Quantitative Results for Layer-Wise Representation}
To assess semantic layer-wise representation, we provide additional comparisons using the Vector Compactness (VeC)~\cite{wang2025layered} and CLIPScore~\cite{hessel2021clipscore} metrics  on the Face and Scene datasets, each containing 100 images. 

VeC assesses how well the primitives are contained within the semantic object boundaries.
VeC is computed by randomly sampling four semantic masks per image and measuring vectors with over 85\% area overlap. 
Table~\ref{tab:vec_results} shows our method achieves higher VeC scores, indicating better containment of primitives within semantic object boundaries.

\begin{table}[htbp]
  \caption{VeC (\%) comparison on Face and Scene datasets.}
  \label{tab:vec_results}
  \centering
  \small
  \setlength{\tabcolsep}{4pt}
  \renewcommand{\arraystretch}{0.85}
  \begin{tabular}{lcccccc}
    \toprule
    Dataset & VeC (\%)
    & DiffVG 
    & LIVE 
    & O\&R 
    & LayerVec 
    & Ours \\
    \midrule
    \multirow{2}{*}{Face}
    & Avg.
    & 48.3
    & 51.9
    & 49.3
    & \underline{55.4}
    & \textbf{65.2$\uparrow$} \\
    & Std.
    & 11.7
    & 11.5
    & 12.3
    & 10.8
    & 8.6 \\
    \midrule
    \multirow{2}{*}{Scene}
    & Avg.
    & 53.4
    & 55.6
    & 52.8
    & \underline{57.6}
    & \textbf{66.2$\uparrow$} \\
    & Std.
    & 12.8
    & 12.2
    & 12.8
    & 11.1
    & 9.2 \\
    \bottomrule
  \end{tabular}
\end{table}

\begin{figure}[htbp]
  \centering
   \includegraphics[width=0.99\linewidth]{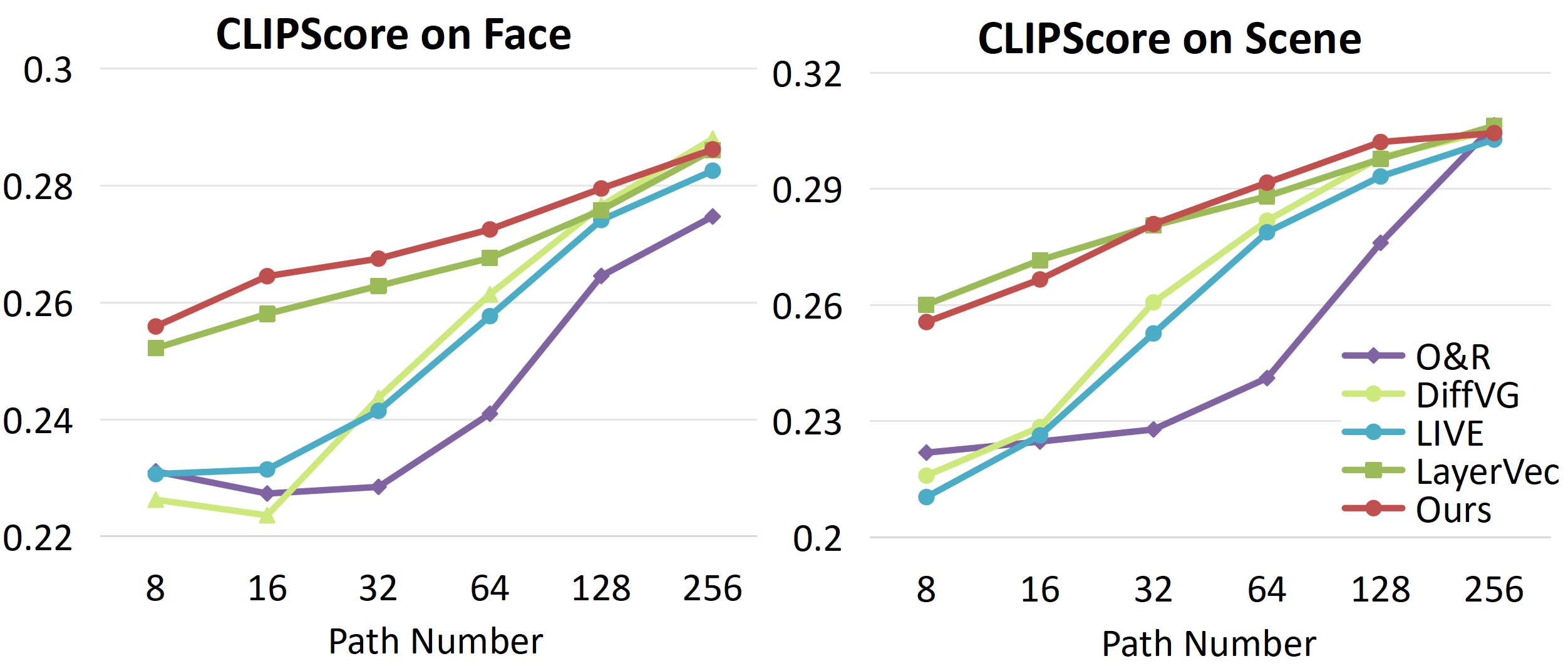}
   \caption{Comparison of CLIP similarity score.}
   \label{fig:clipscore}
\end{figure}

To quantify semantic alignment between the rendered SVG images and the original image content, we use CLIPScore as a metric of cross-modal semantic similarity. We first generate captions for each target image using BLIP~\cite{li2022blip}, and then compute the CLIP similarity between the rendered SVG images and the corresponding target captions. 
In Figure~\ref{fig:clipscore}, our method achieves competitive CLIP scores across path budgets, demonstrating stable layer-wise semantic alignment.

\section{Blending Mode of SVG}

Given the decomposition in Eq.~\ref{eq: decompose}, the albedo, shade, and light
layers must be composited into a unified SVG representation to produce the final
output. We 
use the native blending modes defined in the SVG specification to implement the
image formation model:
\[
\mathcal{V}_{\text{final}}
= \mathcal{V}_{A} * \mathcal{V}_{S} + \mathcal{V}_{L}.
\]
Here, the element-wise multiplication between the albedo and shade layers is
implemented with the \texttt{multiply} blending mode, while the additive
light term is implemented using the \texttt{plus-lighter} blending mode.

The \texttt{multiply} mode performs channel-wise multiplication between the two overlapping layers, producing darker tones in intersecting regions.
This behavior closely matches the attenuation effect of shading in intrinsic
image decomposition, allowing the shade layer to modulate surface reflectance
without altering the underlying albedo structure. In contrast, the
\texttt{plus-lighter} mode performs linear additive blending and is therefore
well aligned with modeling highlights or residual illumination. This operator
preserves high-frequency lighting cues and naturally accumulates light
intensities in overlapping areas.

\begin{figure}[htbp]
  \centering
  \begin{subfigure}[b]{0.49\linewidth}
    \centering
    \includegraphics[width=\linewidth]{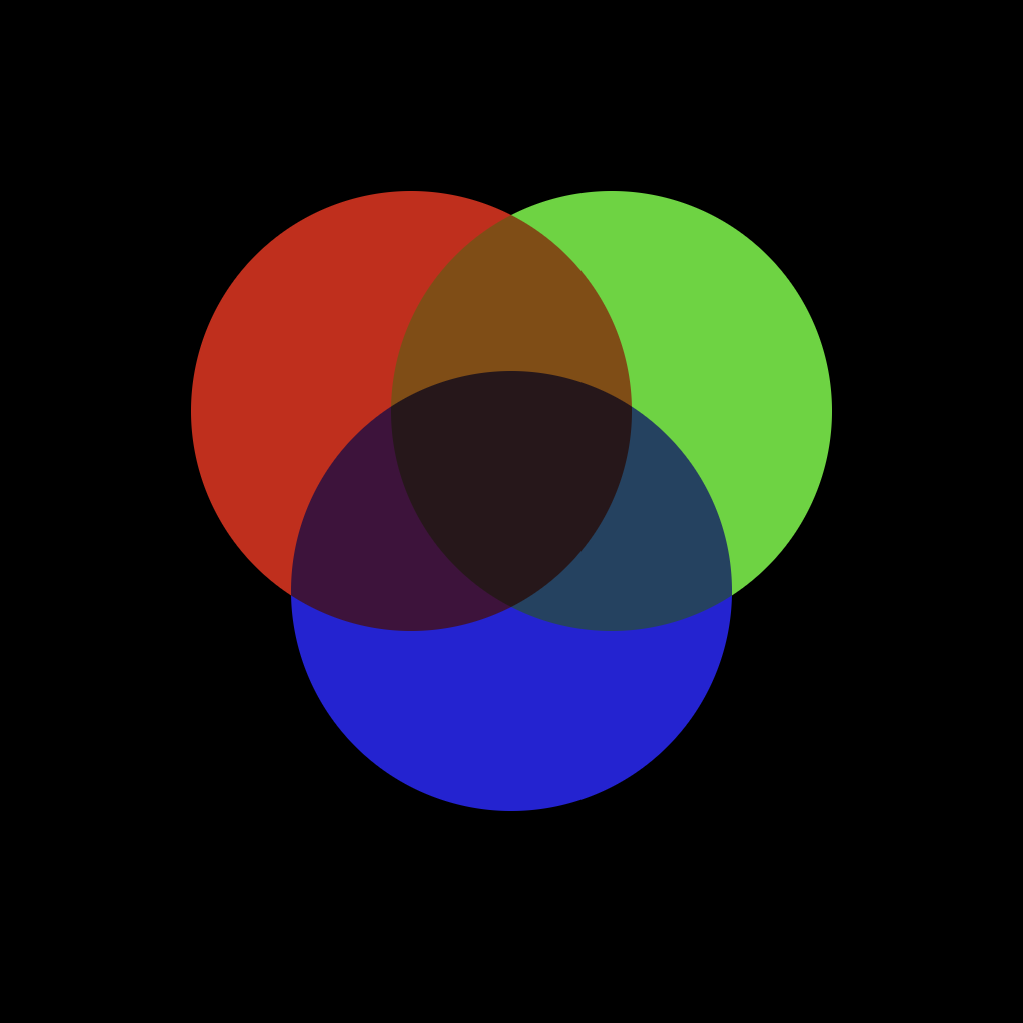}
    \caption{Multiply Blending.}
    \label{fig:short-a}
  \end{subfigure}
  \hfill
  \begin{subfigure}[b]{0.49\linewidth}
    \centering
    \includegraphics[width=\linewidth]{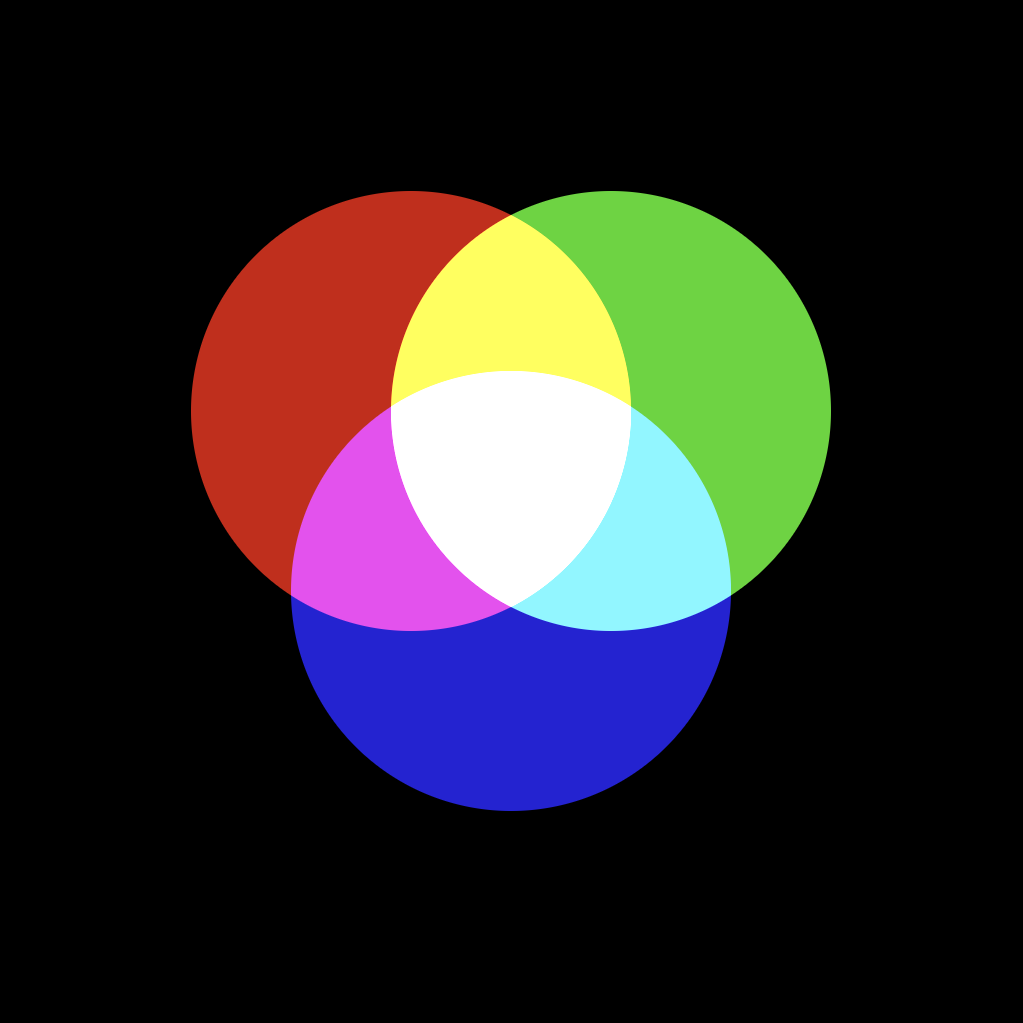}
    \caption{Plus-lighter Blending.}
    \label{fig:short-b}
  \end{subfigure}
  \caption{Illustration of different SVG blending modes: multiply and plus-lighter.}
  \label{fig:short}
\end{figure}

Figure~\ref{fig:short} provides a visual illustration of these two blending
modes. By leveraging SVG's native compositing operators, our method achieves a
fully vectorized representation whose rendering behavior directly corresponds to
the intrinsic image formation model, while maintaining resolution-independence,
editability, and semantic interpretability of individual layers.

\section{Details of Layer Separation}
After illumination refinement, the illumination layer $\mathcal{V}_{I}$ is separated into the shade layer $\mathcal{V}_{S}$ and the light layer $\mathcal{V}_{L}$. 
The separation is performed by examining each vector path’s fill color intensity. 
Let each path in $\mathcal{V}_{I}$ be represented as
\[
\theta_n^{I} = \{P_n^{I},\, C_n^{I},\, \tau_n^{I}\},
\]
where $P_n^{I}$ denotes the geometric shape, $C_n^{I} \in \mathbb{R}^3$ is the RGB fill color, and $\tau_n^{I}$ is the opacity. 
A path is assigned to either the shade or light layer according to the normalized intensity condition:
\[
\begin{aligned}
\theta_n^{S} &= 
\begin{cases}
\theta_n^{I}, & \|C_n^{I}\|_{\infty}\le 1,\\
\varnothing, & \text{otherwise},
\end{cases} \\[4pt]
\theta_n^{L} &= 
\begin{cases}
\varnothing, & \|C_n^{I}\|_{\infty}\le 1,\\
\{P_n^{I}, C_n^{L}, 1\}, & \text{otherwise},
\end{cases}
\end{aligned}
\]
where $\|C_n^{I}\|_{\infty} = \max(C_n^{I})$ measures per-path color intensity.
Paths whose colors remain in the normalized range $[0,1]$ are therefore grouped into $\mathcal{V}_S$, while those exceeding this range are treated as highlight-contributing and grouped into $\mathcal{V}_L$.
During this process, all control points remain unchanged: shade-layer paths inherit both shapes and colors from $\mathcal{V}_{I}$, whereas light-layer paths inherit only the shapes.

Because the light layer uses additive blending, colors for paths in $\mathcal{V}_{L}$ cannot be taken directly from $\mathcal{V}_{I}$. 
Instead, they are recomputed from the residual illumination, defined as
\[
\mathcal{R}_{\text{res}}
    = \mathcal{I}
    - \mathcal{R}(\mathcal{V}_{A}) * \mathcal{R}(\mathcal{V}_{S}),
\]
where $\mathcal{R}(\cdot)$ denotes differentiable vector rendering.
The final color for each light-layer path is obtained by averaging the residual illumination within its covered region:
\[
C_n^{L} = 
\frac{1}{|P_n^{I}|}
\sum_{x \in P_n^{I}} 
\mathcal{R}_{\text{res}}(x).
\]

Finally, the shade and light layers are given by
\[
\mathcal{V}_{S} = \sum_{n} \theta_n^{S},
\qquad
\mathcal{V}_{L} = \sum_{n} \theta_n^{L}.
\]

\section{Experimental Setup}

\subsection{Implementation Details}
During the structural optimization stage, we perform a total of 100 iterations: the first 50 iterations use only the structural loss to stabilize geometric alignment, while the remaining 50 iterations switch to the reconstruction loss to enhance overall visual fidelity.
During the illumination refinement stage, each newly added batch of illumination paths is optimized for 100 iterations, including both parameter refinement and redundant path pruning. The number of path-addition rounds is capped at 5 to prevent unnecessary path growth and avoid overfitting.

\subsection{Model Variant for simple images.}
For stylized or emoji-like images with minimal illumination variation, we adopt a simplified configuration that omits the shade–light decomposition and optimizes only the albedo layer. At initialization, we perform semantic segmentation on the input image to obtain coarse region masks. These masks are then complemented by those produced through our proposed region-wise semantic binarization strategy. The two sets of masks are concatenated to guide the initialization of vector paths and the subsequent structural optimization. After the structural optimization stage, additional paths are iteratively added and pruned solely within the albedo layer, yielding the final compact SVG representation.

\subsection{Controlled Color-Editing Experiment}

Given an original image $I \in \mathbb{R}^{H \times W \times 3}$, a reference edited image 
$\hat{I} \in \mathbb{R}^{H \times W \times 3}$, and an SVG representation 
$\mathcal{V} = \{\,\theta_n\,\}_{n=1}^{N}$ consisting of $N$ vector paths, our controlled color-editing procedure automatically determines a set of editable paths and assigns new colors consistent with the target modification.

\textbf{Edit Region Detection.}  
To localize regions that differ between $I$ and $\hat{I}$, we compute a per-pixel difference map:
\[
D(p) = \frac{1}{3} \sum_{c=1}^{3} \left| I_c(p) - \hat{I}_c(p) \right|, 
\qquad p \in \Omega,
\]
where $\Omega$ denotes the image domain.  
A binary edit mask $M_{\text{edit}}$ is then obtained via thresholding:
\[
M_{\text{edit}}(p) = 
\begin{cases}
1, & D(p) > \tau, \\
0, & \text{otherwise},
\end{cases}
\]
with $\tau$ a fixed scalar threshold.

\textbf{Per-Path Masking and Candidate Selection.}
Each vector path $\theta_n$ is rendered independently to obtain a binary mask 
$M_n \in \{0,1\}^{H \times W}$.  
We measure how well a path spatially overlaps with the edit region using IoU:
\[
\text{IoU}(n) 
= \frac{\sum_{p} M_n(p) M_{\text{edit}}(p)}
       {\sum_{p} \max(M_n(p),\, M_{\text{edit}}(p))}.
\]
Paths with $\text{IoU}(n) > \gamma$ (a small constant) are retained as spatially relevant candidates.

We further enforce color compatibility.  
Let
\[
\mu_n = \frac{1}{|S_n|}\sum_{p \in S_n} I(p),
\qquad
\hat{\mu}_n = \frac{1}{|S_n|}\sum_{p \in S_n} \hat{I}(p),
\]
where $S_n = \{ p : M_n(p)=1 \}$ is the support of path $n$.  
A path is kept only if 
\[
\left\| \mu_n - \hat{\mu}_n \right\|_2 \le \delta,
\]
ensuring that the path corresponds to a region that actually changes color in the edited image.

Among the retained paths, we sort them by $|S_n|$ (descending) and choose the top 
$K \in \{1,2,4,8,16\}$ paths:
\[
\mathcal{S}_K = \operatorname*{TopK}_{n} |S_n|.
\]

\textbf{Automatic Color Assignment.}
For each selected path $n \in \mathcal{S}_K$, we compute the target color:
\[
C^{\text{ref}}_n 
= \frac{1}{|S_n|}\sum_{p \in S_n} \hat{I}(p).
\]

If the vectorization method includes an illumination model with a shade image 
$S(p)$ (e.g., $\mathcal{V}_S$), we compute the mean shade in the same region:
\[
\bar{S}_n = \frac{1}{|S_n|}\sum_{p \in S_n} S(p),
\]
and derive the updated albedo color via
\[
C^{A}_n = \frac{C^{\text{ref}}_n}{\bar{S}_n + \varepsilon},
\]
where $\varepsilon$ avoids division by zero.

For methods without illumination decomposition, we directly set
\[
C^{A}_n = C^{\text{ref}}_n.
\]

The fill color of path $n$ is then replaced by $C^{A}_n$, producing an updated SVG:
\[
\mathcal{V}' = \{\,\theta'_n\,|\, n \in \mathcal{S}_K\ \text{updated}, \ 
n \notin \mathcal{S}_K\ \text{unchanged}\}.
\]


\begin{figure*}[htbp]
  \centering
   \includegraphics[width=0.65\linewidth]{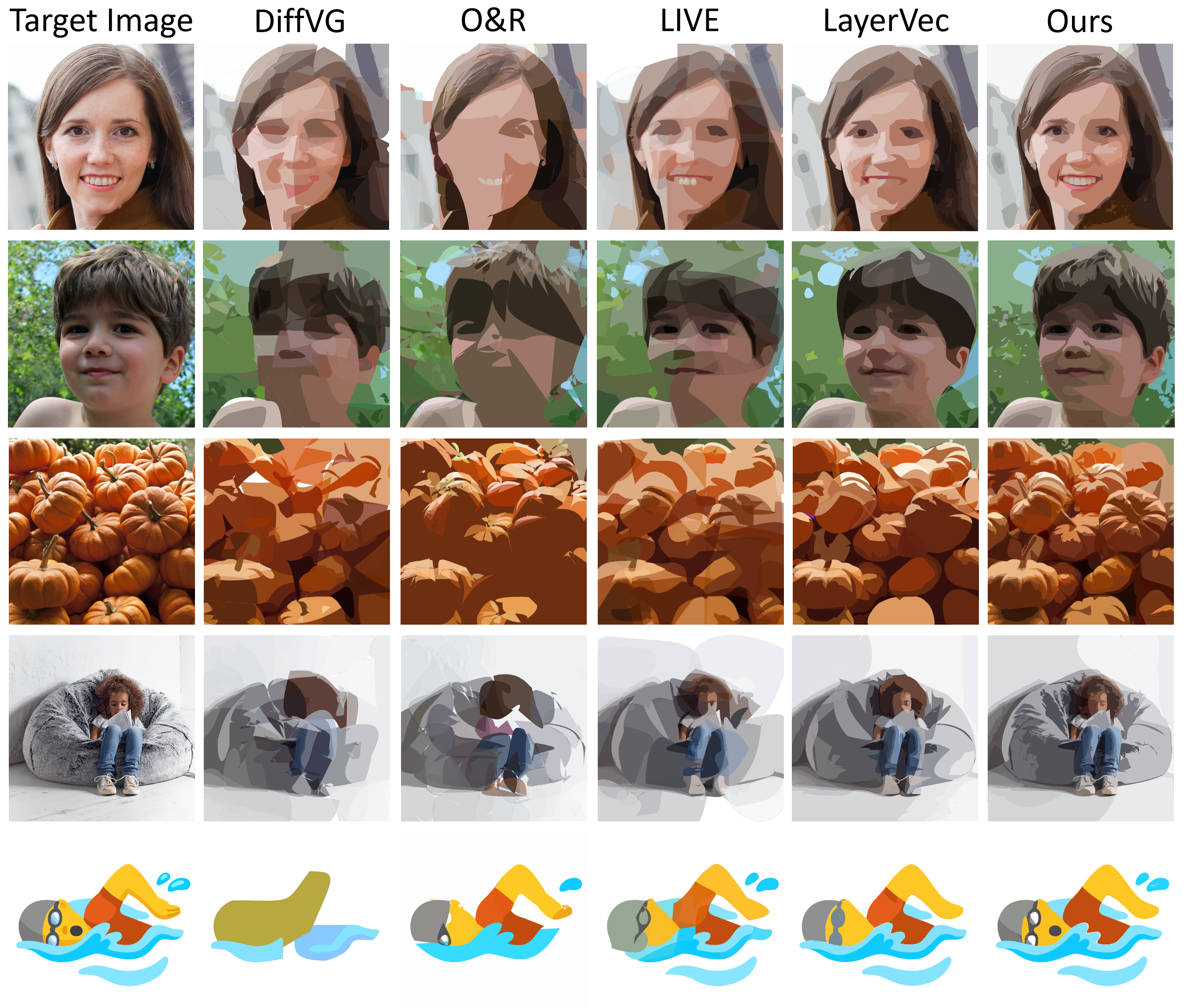}
   \caption{Qualitative reconstruction comparison across datasets. All examples in the Face and Scene datasets are vectorized using 64 paths, whereas Emoji examples are vectorized with 16 paths.}
   \label{fig:fide_suppl}
   \vspace{-0.3em}
\end{figure*} 

\begin{figure*}[htbp]
  \centering
  \begin{subfigure}[b]{0.45\linewidth}
    \centering
    \includegraphics[width=\linewidth]{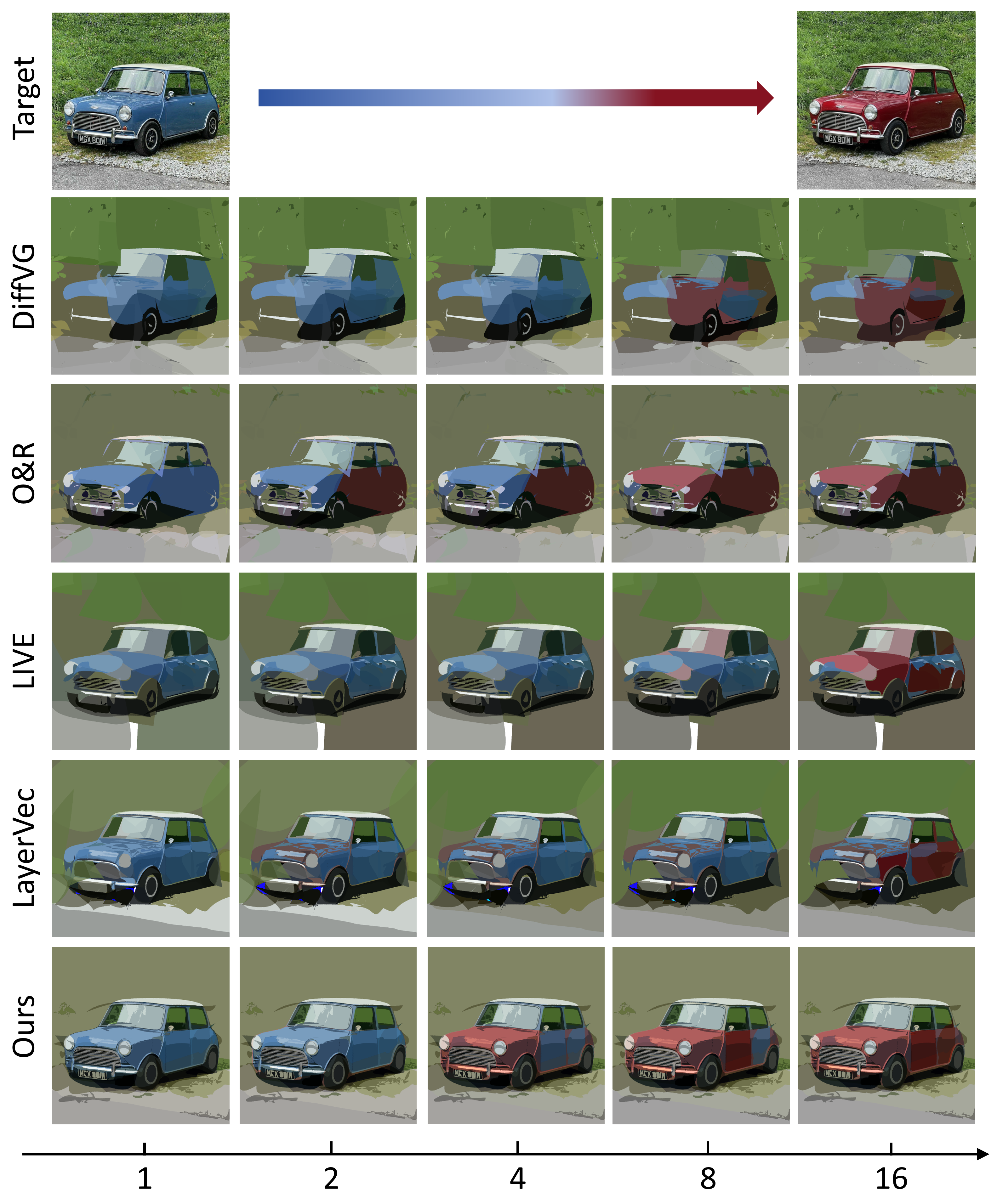}
    \caption{ }
    \label{fig:edit_suppl_1}
  \end{subfigure}
  \hfill
  \begin{subfigure}[b]{0.45\linewidth}
    \centering
    \includegraphics[width=\linewidth]{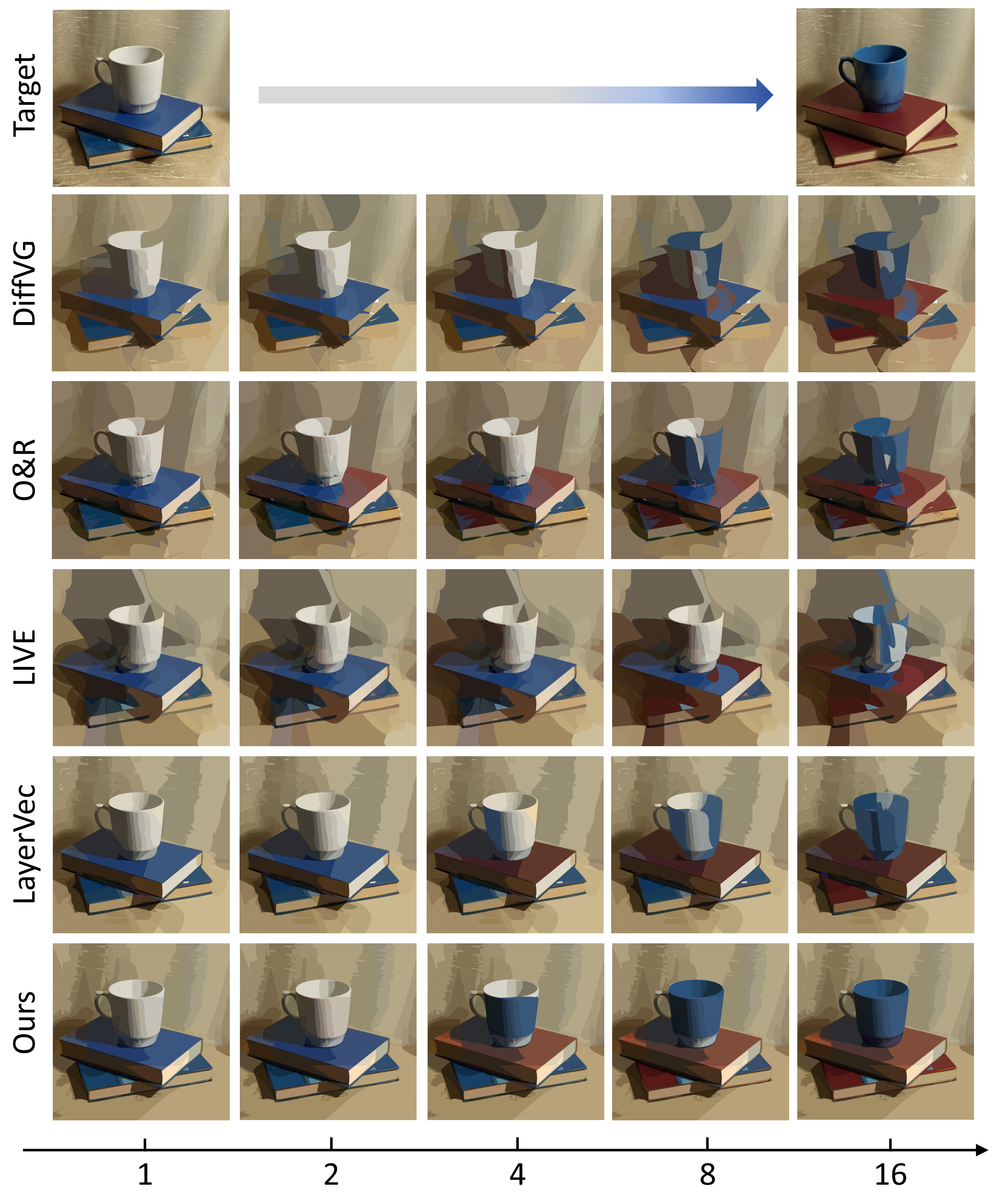}
    \caption{ }
    \label{fig:edit_suppl_2}
  \end{subfigure}
  \caption{Color-editing comparison with controlled path modification.
  Each column corresponds to the number of edited paths.}
  \label{fig:edit_suppl}
\end{figure*}

\section{Additional Qualitative Results}
\begin{figure*}[htbp]
  \centering
  \begin{subfigure}[b]{0.7\linewidth}
    \centering
    \includegraphics[width=\linewidth]{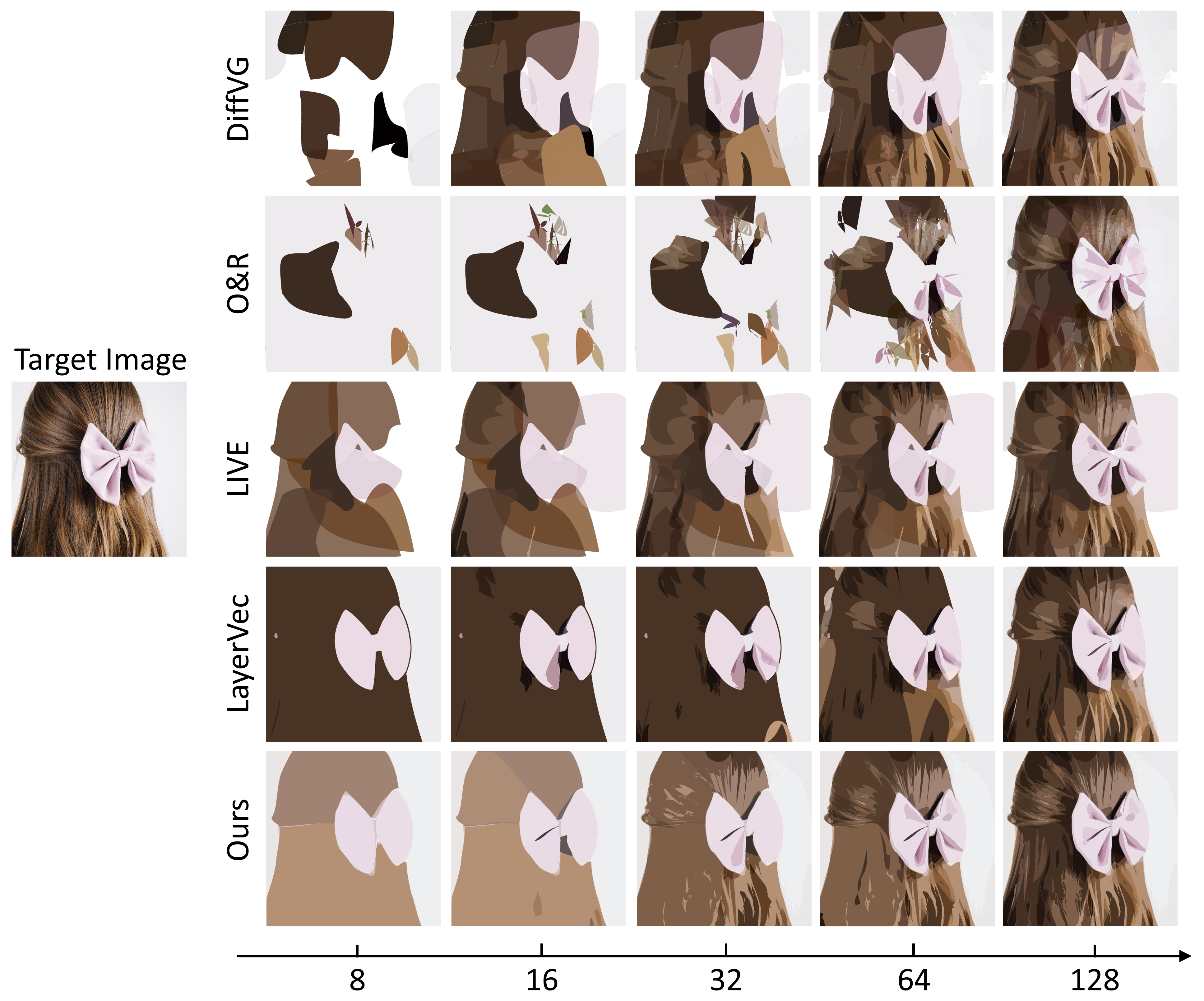}
    \caption{ }
    \label{fig:layer_suppl_1}
  \end{subfigure}
  \hfill
  \begin{subfigure}[b]{0.7\linewidth}
    \centering
    \includegraphics[width=\linewidth]{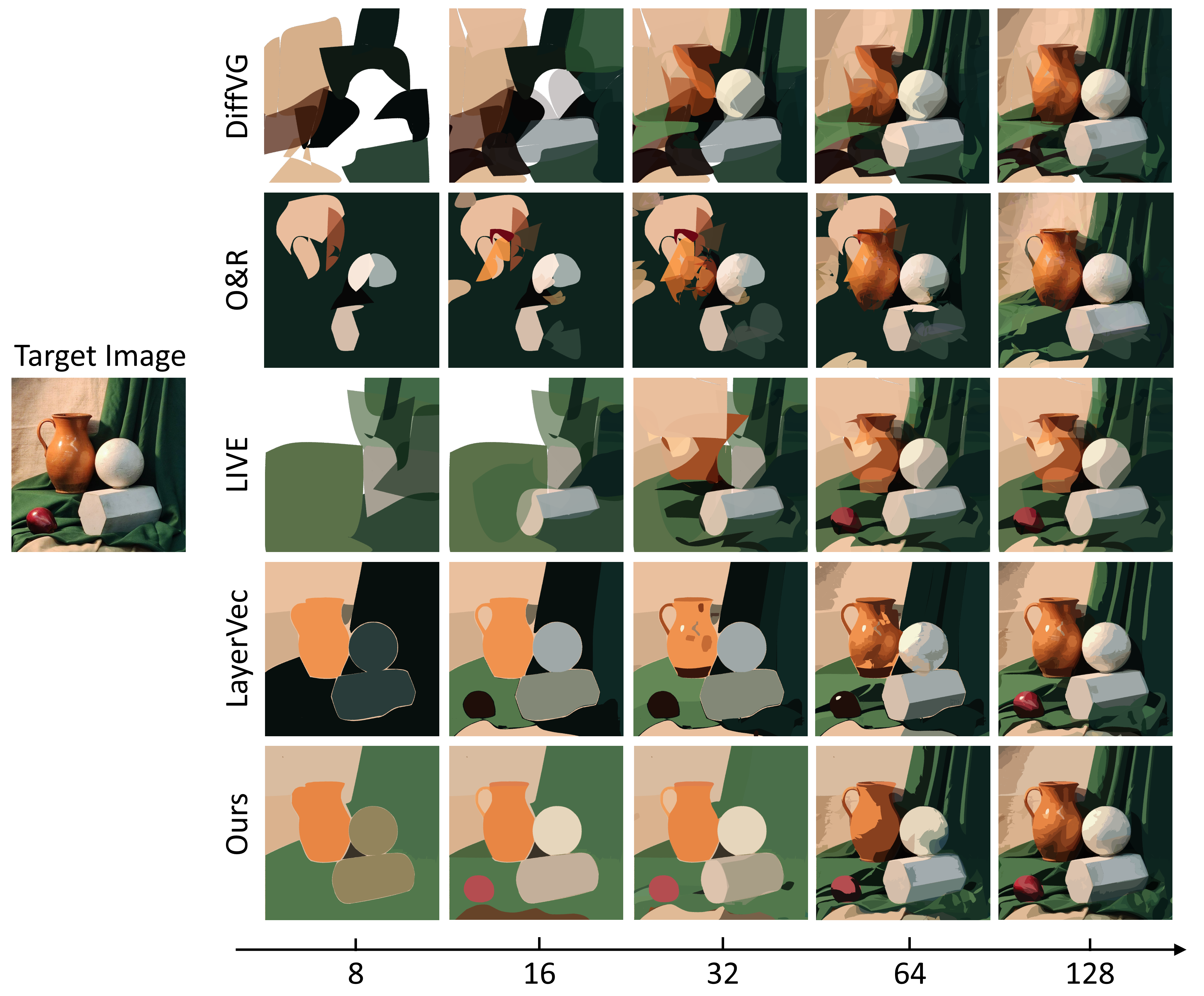}
    \caption{ }
    \label{fig:layer_suppl_2}
  \end{subfigure}
  \caption{Comparison of layer-wise representation.
  Each column shows the result with an increasing number of paths, revealing how the layered representation is progressively organized.}
  \label{fig:layer_suppl}
\end{figure*}

\begin{figure*}[htbp]
  \centering
  \begin{subfigure}[b]{0.7\linewidth}
    \centering
    \includegraphics[width=\linewidth]{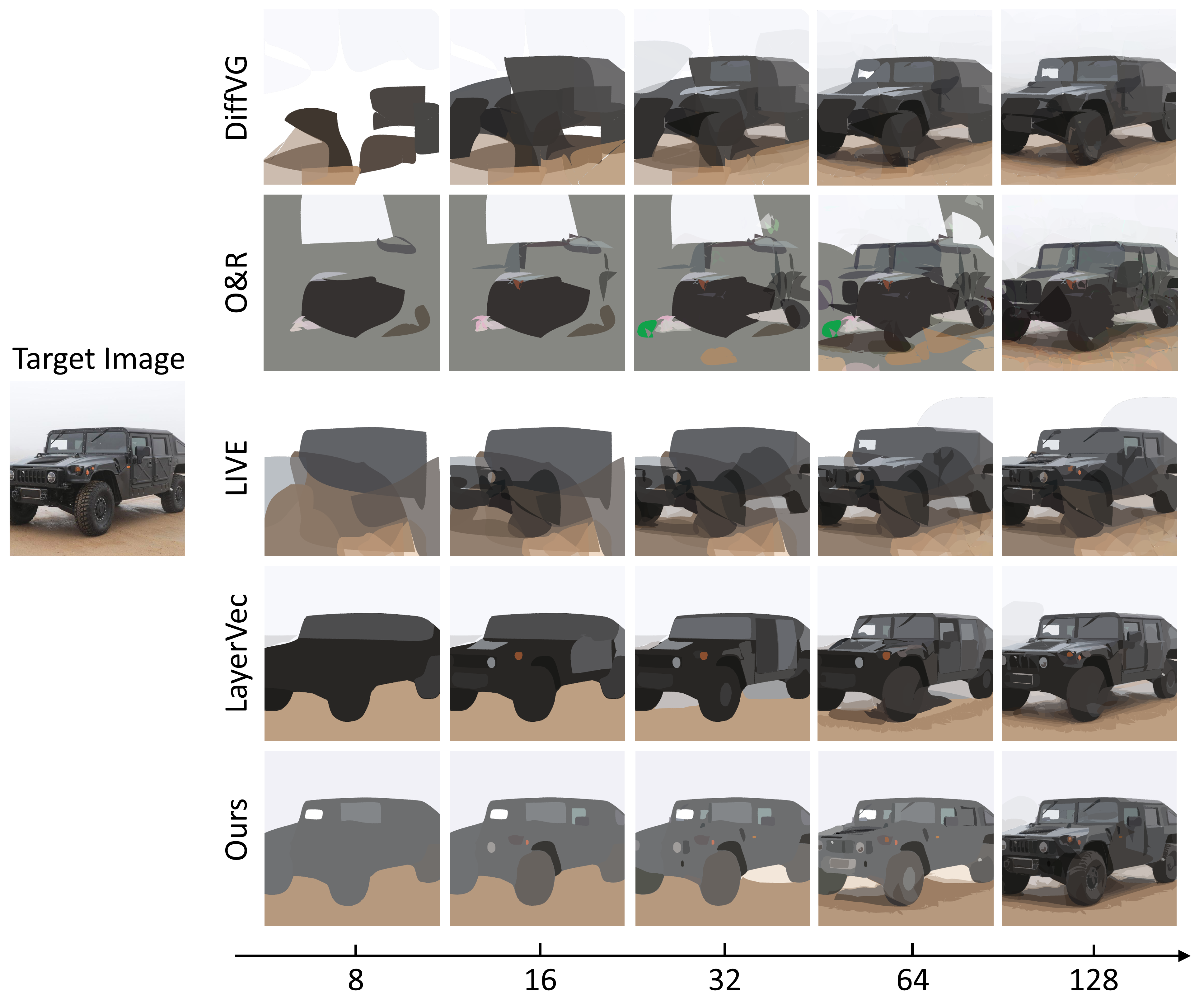}
    \caption{ }
    \label{fig:layer_suppl_11}
  \end{subfigure}
  \hfill
  \begin{subfigure}[b]{0.7\linewidth}
    \centering
    \includegraphics[width=\linewidth]{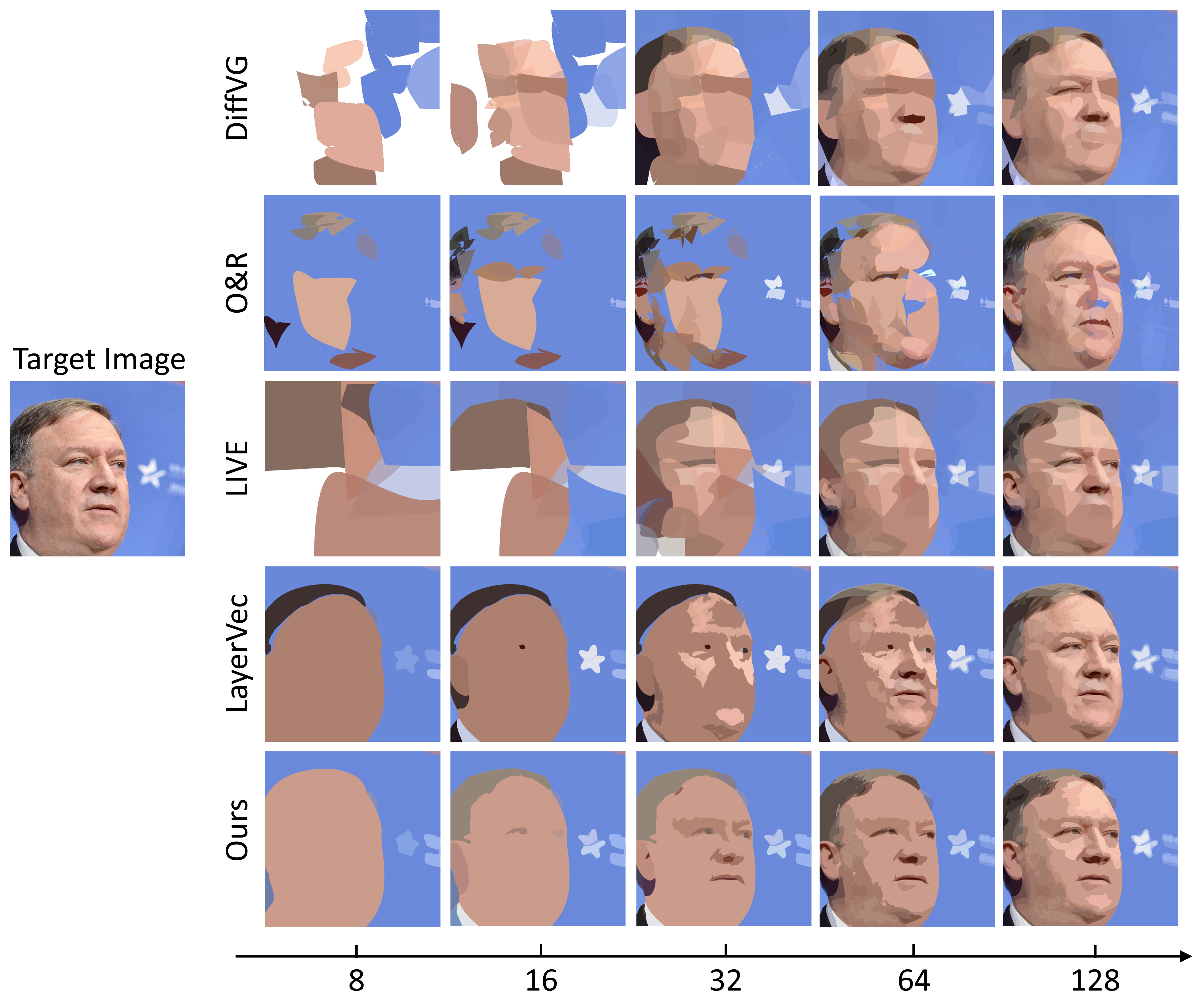}
    \caption{ }
    \label{fig:layer_suppl_12}
  \end{subfigure}
  \caption{Comparison of layer-wise representation.
  Each column shows the result with an increasing number of paths, revealing how the layered representation is progressively organized.}
  \label{fig:layer_suppl1}
\end{figure*}

We compare COVec with four representative vectorization methods. Figure~\ref{fig:fide_suppl} shows additional reconstruction results, where our method preserves both structure and tonal detail under the same path budget.
Figure~\ref{fig:edit_suppl} illustrates the editability of the SVG produced by our method.
Figures~\ref{fig:layer_suppl} and \ref{fig:layer_suppl1} further visualize our layer-wise representation.


\end{document}